\documentclass[11pt]{article}

\usepackage{acl}

\usepackage{times}
\usepackage{latexsym}

\usepackage[T1]{fontenc}

\usepackage[utf8]{inputenc}

\usepackage{microtype}

\usepackage{inconsolata}

\usepackage{graphicx}

\usepackage{booktabs}
\usepackage{arydshln}

\definecolor{stdgray}{gray}{0.4} 
\newcommand{\stdev}[1]{{\tiny\textcolor{stdgray}{$\pm#1$}}} 

\title{``Act Like a 5th Grader'' is Not Enough: Bounding Knowledge in LLM-Based User Simulators}

\author{Krisztian Balog \qquad Arild Michel Bakken \\
        University of Stavanger, Norway \\
        \texttt{\{krisztian.balog,arild.m.bakken\}@uis.no}
      }

\begin{document}
\maketitle
\begin{abstract}
  Large language models (LLMs) are increasingly used to simulate human behavior but frequently fail to exhibit realistic cognitive constraints, suffering from a ``superhuman bias.'' Using a dataset of over 71,000 reading comprehension responses from 2,359 primary-school students (grades 4--6), we demonstrate that standard persona prompting yields near-perfect, deterministic performance, failing to capture the natural variance of developing readers. To address this, we introduce the Cognitively Bounded User Simulator (CBUS), an architectural framework that explicitly models the restricted working memory of young readers through an episodic bottleneck. Within this framework, we formalize two distinct test-taking strategies to emulate different reading behaviors. Our evaluation shows that explicitly modeling cognitive bounds significantly narrows the simulation gap across multiple LLM backbones, demonstrating that enforcing architectural constraints is more effective for high-fidelity simulation than simply scaling raw model capabilities.
\end{abstract}

\section{Introduction}

Large language models (LLMs) excel at ``persona playing,'' making them a convenient tool to simulate human behavior in various application contexts, including the automatic evaluation of conversational agents~\citep{yoon-etal-2024-evaluating}, counseling~\citep{yang-etal-2025-consistent}, and social science experiments~\citep{Argyle:2023:PA,Park:2023:UIST}. By conditioning an LLM on a specific persona or goal, researchers can approximate human interactions and prototype interactive systems at a fraction of the cost of large-scale human evaluation. However, the utility of these simulators hinges entirely on their fidelity to authentic human behavior.

Despite their widespread use, a fundamental limitation of current LLM-based simulators is their inconsistency with authentic human behavior. Simulators frequently confuse their assigned roles, fail to adhere to specified personas, and exhibit a ``superhuman bias'' by displaying omniscient global knowledge and reasoning capabilities far beyond their intended character~\citep{Wang:2024:WWW,Naous:2026:ICLR,Zhou:2026:COLM}.
We argue that these limitations originate from a critical missing component: current LLM-based user simulators rely predominantly on superficial prompting and lack explicit modeling of the underlying human cognitive processes. Crucially, isolating and quantifying these cognitive failures in complex, open-ended simulation environments (such as interacting with conversational agents) is methodologically challenging. When a simulation fails to mimic human limitations, it is difficult to determine whether the cause is a poorly defined persona, shifting context dynamics, or a fundamental architectural inability to bound knowledge.

\begin{figure}[t]
    \centering
    \includegraphics[width=\linewidth]{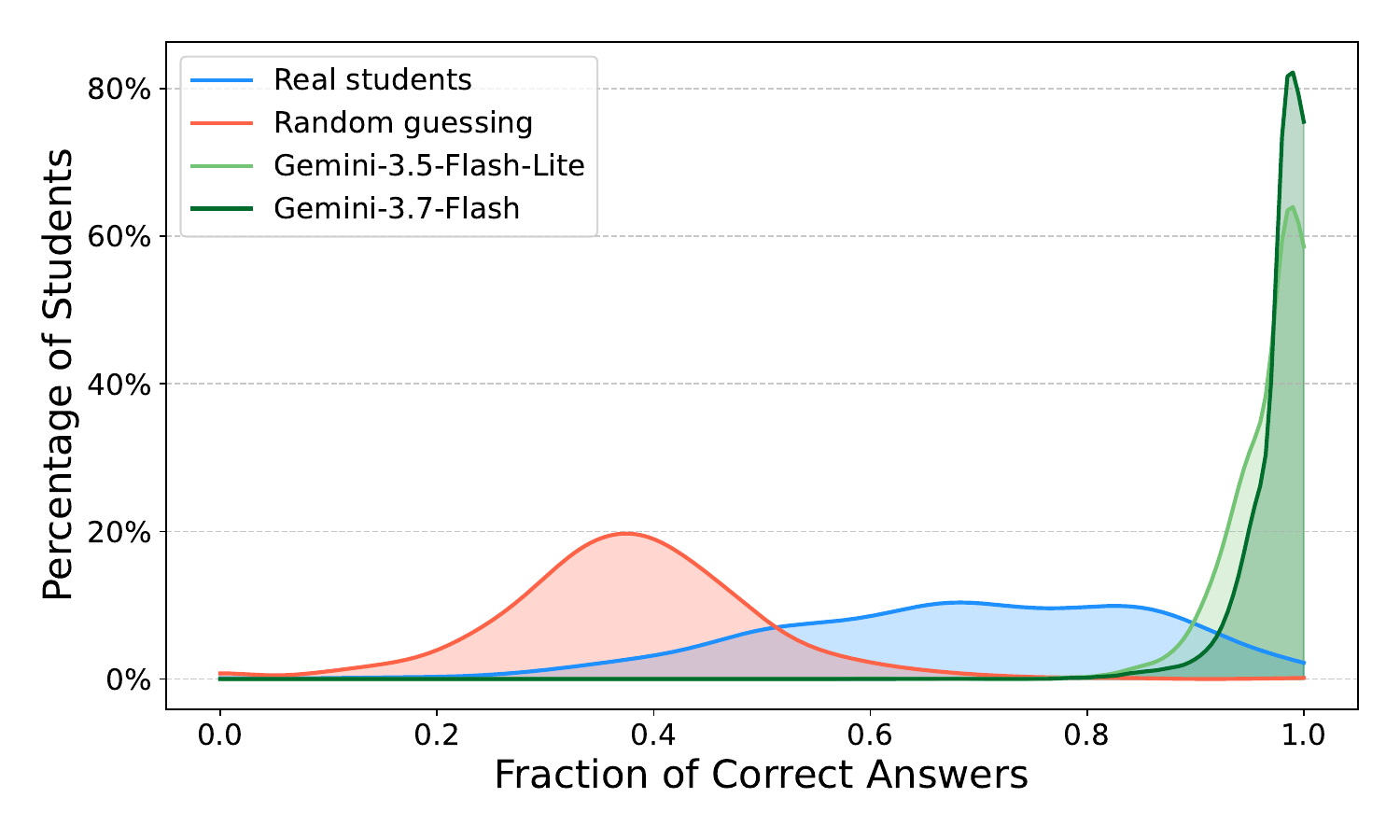}
    \caption{Distribution of per-student scores for real students, random guessing, and two baseline persona-prompted frontier LLMs. Both Gemini models collapse into a near-perfect ``superhuman'' spike that misses the natural performance spread of real students; moreover, the more capable 3.7-Flash model is \emph{more} concentrated than 3.5-Flash-Lite, showing that greater model capability does not translate into more realistic simulation.}
    \label{fig:illustration}
\end{figure}

To rigorously study the challenge of simulating bounded human knowledge, we argue that the evaluation environment must be tightly constrained. We propose standard reading comprehension tests as an ideal, controlled testbed for this type of simulation. By simulating a primary school student taking a standardized reading assessment, we eliminate the variables and confounding factors of open-ended environments. Instead, the simulator's knowledge and cognitive load are strictly bounded by two transparent, verifiable factors: a specific source text and a defined developmental stage.

Despite the constraints of this environment, we show that standard persona-based prompting is highly insufficient for producing realistic simulations. We leverage a unique, large-scale dataset comprising over 71,000 responses from 2,359 students in grades 4--6 to 156 reading comprehension texts and 750 associated questions from the Norwegian educational system. As illustrated in Figure~\ref{fig:illustration}, simply prompting an LLM to ``act like a 5th-grade student'' results in a drastic overestimation of student capabilities. LLMs fail to behave like a developing human reader, effortlessly succeeding at inferential reasoning and vocabulary tasks that demonstrably challenge real students.

To bridge this gap, we argue that high-fidelity simulation requires moving beyond superficial persona assignment and toward explicit cognitive modeling. Rather than simply instructing an LLM to be a student, we must architecturally force it to process information like one. In this paper, we introduce the Cognitively Bounded User Simulator (CBUS), a novel framework grounded in the well-established finding that working memory capacity constrains developing readers' comprehension~\citep{Cain:2004:JEP,Cowan:2001:BBS}. Specifically, we impose a generic working-memory capacity bottleneck through a constrained, two-stage reasoning pipeline: an encoding stage that strictly limits the number of text propositions the simulator can extract, followed by an execution stage where the model is forced to answer comprehension questions utilizing only this restricted memory trace.

By grounding our simulator in these verifiable cognitive processes, our proposed modeling successfully reduces the gap between real students and AI simulations. Our main contributions are fourfold:
\begin{itemize}
    \item We expose and quantify the ``superhuman bias'' of LLM simulators in a highly constrained evaluation environment, demonstrating the inadequacy of standard persona prompting.
    \item We propose CBUS, an architectural framework that effectively models the restricted working memory of a young reader via a parameterized, two-stage episodic bottleneck under two distinct test-taking strategies.
    \item We propose a suite of evaluation metrics specifically tailored for the reading comprehension simulation scenario, moving beyond aggregate accuracy to holistically assess both student-centric and item-centric alignment.
    \item We provide robust empirical validation of our approach using a massive, real-world dataset, establishing that explicit cognitive bounds successfully reduce the calibration gap between simulated and real student responses.
\end{itemize}
Code, data, and prompts are made publicly available at \url{https://github.com/iai-group/emnlp2026-simreading}.
\section{Related work}
\label{sec:related}

User simulation has seen a surge of interest across multiple domains, emerging as a critical methodology for both evaluating and training interactive systems. Beyond their established role in iterating on search, recommendation, and information access systems~\citep{Balog:2024:FnTIR,Balog:2025:SIGIRb}, simulators are increasingly utilized as scalable proxies for human subjects in broad social science studies~\citep{Argyle:2023:PA,Park:2023:UIST,Piao:2026:iFuture}.

Currently, the predominant approach to modern user simulation relies heavily on LLMs, typically achieved through either in-context learning or task-specific fine-tuning~\citep{Terragni:2023:arXiv,kolluri-etal-2025-finetuning,meshi-etal-2026-convapparel}. However, while highly capable, LLM-based simulations frequently fall short of true human fidelity. Because LLMs are optimized to be ``helpful assistants'' rather than cognitively plausible humans, current LLM-based simulators face severe limitations: they systematically diverge from actual human interaction patterns~\citep{yoon-etal-2024-evaluating,seshadri-etal-2026-lost}, exhibit ``superhuman'' knowledge~\citep{Wang:2024:WWW}, and fail to capture the cognitive diversity inherent in real populations~\citep{Davidson:2023:arXiv,Wang:2024:WWW}. Merely prompting these systems to ``act human'' yields poor results~\citep{Naous:2026:ICLR}, as they inherently struggle to adhere to knowledge constraints and instructions~\citep{Kiesel:2024:ECIR,Wang:2024:WWW}.

To address these shortcomings, recent studies have revealed the necessity of moving away from the assumption of ``perfect users.'' Instead, there is a growing push toward explicitly modeling human-like behavior to create more robust and realistic evaluation environments. To bridge the simulation gap, researchers have begun prompting models with explicit user profiles or personas~\citep{wang-etal-2025-know}, tracking users' internal states~\citep{Wu:2026:ICML}, and instilling distinct behavioral traits~\citep{ferreira-etal-2024-multi}. Further advancements involve modeling underlying cognitive processes---such as evolving knowledge states and shifting information needs~\citep{Zerhoudi:2024:JCDL,Zhang:2025:SIGIR}---as well as capturing the realities of non-collaborative user behavior~\citep{Shim:2026:ICLR}.

The need for such bounded simulators is particularly acute in educational settings. Adaptive assessments, which adjust tasks based on performance, have existed for decades but remain limited by the need for expensive empirical calibration via human pretests~\citep{Wainer:2000:Bookchapter}. There is, therefore, a significant demand for high-fidelity simulations that can predict how different student profiles interact with educational content~\citep{vonDavier:2024:Bookchapter}. While recent work uses LLMs to automate this calibration~\citep{yancey-etal-2024-bert}, these models often fail to capture the specific cognitive boundaries of young learners. Our work directly bridges this gap by enforcing structural cognitive limits on the LLM.

A parallel line of work uses LLMs and generative agents to simulate learners directly. EduAgent~\citep{Xu:2024:arXiv} models a single student's fine-grained, within-lecture behavior (gaze, motor actions, cognitive-state transitions) over time. Agent4Edu~\citep{Gao:2025:AAAI} generates learner response data for adaptive tutoring pipelines, initializing agent profiles from the corresponding learner's interaction histories and cognitive traits (problem-solving ability and knowledge proficiency). \citet{Sonkar:2024:arXiv} fine-tune LLMs to reproduce specific misconceptions, namely common algebraic errors of secondary school students. \citet{zelikman-etal-2023-generating} fine-tune LLMs to simulate K-12 students' item responses in order to generate and psychometrically calibrate reading-efficiency tests. Closest to our setting, Generative Students~\citep{Xiao:2024:LAS} answer multiple-choice items using hand-specified knowledge-component profiles to support item evaluation. Our work differs from all of the above on three axes: (i) we require no interaction data, knowledge-component tagging, or misconception taxonomy---inputs unavailable for a one-shot standardized reading test; (ii) our target is \emph{population-level distributional fidelity}---matching the score and per-item difficulty distributions of real students---rather than individual-learner behavior, item evaluation, or misconception replication; and (iii) our mechanism is a working-memory capacity bottleneck, motivated by cognitive limitations of young learners---which none of these works model.

\section{Reading Comprehension Dataset}
\label{sec:expsetup}

Our dataset comprises (1) a corpus of texts with associated comprehension questions and (2) a collection of responses from a large-scale study involving 2,359 students. All materials and responses are in Norwegian (Bokmål), the primary written standard in Norwegian schools. Our focus on this language is motivated by the availability of high-quality assessment data to study the challenges of LLMs simulating young readers.

\subsection{Texts and Questions}

The dataset contains 156 texts designed to evaluate various dimensions of reading proficiency. Each text is accompanied by a set of questions, which can be of the following types: \emph{multiple choice} (single-select), \emph{true-or-false}, \emph{checkboxes} (multi-select), and \emph{free text}. For the purposes of this study, we exclude free-text responses to focus on objective, automatically gradable question types. This results in a final subset of 750 questions.

The questions are further divided into three categories based on the Norwegian Framework for Basic Skills~\citep{udir_framework_reading}, 
which is closely related to the framework used in the PIRLS international assessment~\citep{Sabatini:2024:PIRLS}. The categories include: \emph{Locate} (finding explicitly or implicitly expressed information), \emph{Interpret \& Connect} (drawing cross-text inferences), and \emph{Reflect \& Evaluate} (engaging critically and independently with texts). This categorization was created semi-automatically using an ensemble of LLM-based classifiers followed by expert review; the full methodology is detailed in Appendix~\ref{app:question_categories}. Table~\ref{tbl:questions} provides a breakdown of the question types and categories.
The corpus of texts and questions is made publicly available. 


\begin{table}[t]
    \centering
    \caption{Question statistics.}
    \label{tbl:questions}
    \small
    \begin{tabular}{l@{}rrr|r}
        \toprule
                    & Locate \& & Interpret \& & Reflect & \textbf{Total} \\
                    & Connect   & Evaluate     &                          \\
        \midrule
        multiChoice & 310       & 40           & 12      & 362            \\
        trueOrFalse & 317       & 23           & 4       & 344            \\
        checkboxes  & 41        & 3            & 0       & 44             \\
        \midrule
        Total       & 668       & 66           & 16      & 750            \\
        \bottomrule
    \end{tabular}
\end{table}

\subsection{Participants and Ethics}
The study utilizes data from 4th to 6th-grade classes (ages 8--11) within the Norwegian primary education system. Data collection followed the ethical guidelines of the National Committee for Research Ethics in the Social Sciences and the Humanities~\citep{NESH2021}. We obtained full parental consent and student assent, and we exclusively analyze data for which this consent was secured. Critically, the process was privacy-preserving by default: no personally identifiable information (PII) was recorded, ensuring all responses remain strictly anonymous. Further administrative details regarding school recruitment, session structure, and exemptions from formal agency evaluation due to data anonymity are delegated to Appendix~\ref{app:ethics}. Throughout, our simulators target a prototypical 5th-grade (10--11 year-old) reader as the representative midpoint of this 4th--6th-grade range.

\subsection{Responses}

Data collection followed a self-selection protocol where students chose texts based on personal interest from randomly distributed pairs. Upon completing a task, students rated the text and were presented with a new pair of randomly selected texts. A screenshot of the user interface and further protocol details are provided in Appendix~\ref{app:dataset}.

\paragraph{Data Formalization}
To standardize the evaluation across heterogeneous question types, we binarize all student inputs into discrete dichotomous responses. For multiple-choice and true-or-false questions, a response is marked as correct (1) if the student selects the single correct answer, and incorrect (0) otherwise. For checkbox questions, we employ strict ``all-or-nothing'' scoring: a response is marked as correct (1) only if the subset of checked boxes perfectly aligns with the gold standard. 
Formally, we define a \emph{response} as a tuple $(s, q) \in \{0,1\}$, representing the binary performance of student $s$ on question $q$.

\paragraph{Dataset Statistics \& Difficulty}
In total, our filtered dataset contains 71,789 responses from 2,359 unique students across 80 classes. Regarding empirical question difficulty, multi-select checkbox questions proved significantly more difficult for the student population than the other two single-select formats. Across pedagogical categories, difficulties remain broadly comparable, though \emph{Interpret \& Connect} questions exhibit slightly lower success rates overall. Additional descriptive statistics are provided in Appendix~\ref{app:dataset}.

\paragraph{Data Splitting}
To rigorously evaluate simulation fidelity and establish an empirical baseline for human variance, we partition the dataset into two disjoint halves. To prevent data leakage (e.g., shared educational environments or teacher effects), this split is performed strictly at the \emph{classroom} level (Table~\ref{tbl:data_split}). Because our simulators operate zero-shot, the conventional ``train/test'' nomenclature is misleading. Instead, we designate one half as the \textbf{Ground Truth} evaluation split, which serves as the primary target distribution for all simulator metrics. The remaining half acts as a \textbf{Held-out} reference split; evaluating this population against the ground truth establishes the natural statistical variance (or ``noise floor'') of real students, providing an empirical upper bound for simulation fidelity.

\begin{table}[t]
    \centering
    \caption{Dataset split statistics.}
    \label{tbl:data_split}
    \small
    \begin{tabular}{l@{}r@{~~~}r@{~~~}r}
        \toprule
        \textbf{Split}            & \textbf{Classes} & \textbf{Responses} & \textbf{Students} \\
        \midrule
        Held-out (Reference)      & 40               & 35,739             & 1,161             \\
        Ground Truth (Evaluation) & 40               & 36,050             & 1,198             \\
        \midrule
        \textbf{Total}            & 80               & 71,789             & 2,359             \\
        \bottomrule
    \end{tabular}
\end{table}

\section{Pilot Study: Evaluating Baseline Simulators}
\label{sec:baseline}

We first investigate whether state-of-the-art LLMs can emulate 5th-graders using standard ``persona-based'' prompting. While it is widely assumed LLMs can adopt personas via prompt engineering, simulating a developing reader requires actively suppressing inherent global knowledge to replicate cognitive limitations. We evaluate baseline zero-shot simulators against empirical student data to quantify the simulation gap and identify where standard prompting fails.

\begin{table}[t]
    \caption{Pilot study results demonstrating the ``superhuman bias'' of baseline persona-based prompting compared to the empirical student distribution. The full metric set with standard deviations is in Appendix~\ref{app:pilot_full}.}
    \label{tab:pilot_results}
    \centering
    \small
    \begin{tabular}{lcc}
        \toprule
        \textbf{Population}          & \textbf{Mean Acc.} & \textbf{JSD ($\downarrow$)} \\
        \midrule
        Real Students (Ground Truth) & 0.687              &                             \\
        Real Students (Held-out)     & 0.661              & 0.006                       \\
        \midrule
        Random Guessing              & 0.378              & 0.518                       \\
        \hdashline
        Llama-3.3-70B-Instruct       & 0.966              & 0.679                       \\
        Mixtral-8x22B-Instruct       & 0.956              & 0.634                       \\
        GPT-4o-mini                  & 0.958              & 0.646                       \\
        GPT-5.4                      & 0.974              & 0.732                       \\
        Gemini-3.5-Flash-Lite        & 0.972              & 0.713                       \\
        Gemini-3.7-Flash             & 0.985              & 0.773                       \\
        \bottomrule
    \end{tabular}
\end{table}

\subsection{Pilot Experimental Setup}

To establish a high-capability baseline, we evaluate zero-shot persona prompting on flagship commercial and open-weight models (version details in Appendix~\ref{app:model_details}) using the template in Figure~\ref{fig:baseline_prompt}. Structural instructions are in English to maximize adherence, while explicitly enforcing a Norwegian persona.\footnote{In Appendix~\ref{app:pilot_full} we test models with Norwegian prompts and show that fully Norwegian instructions stay superhuman, i.e., the gap is not an instruction-language artifact.} Following our primary protocol (Section~\ref{sec:exp_setup}), we set generation temperature to $T=0.7$ and average results over $n=3$ runs.

\subsection{Results and Key Findings}
\label{sec:pilot_results}

We analyze two macro-level metrics: \textbf{Mean Accuracy} (average percentage of correct answers) and \textbf{JSD} (Jensen-Shannon Divergence, comparing individual score distributions between simulated and real Ground Truth populations). To establish an empirical upper bound for simulation fidelity, we also evaluate a held-out split of human students against the Ground Truth.

\paragraph{The Superhuman Bias}
Table~\ref{tab:pilot_results} reveals a stark discrepancy. While the empirical student population exhibits natural variance (mean accuracy 0.687), baseline simulators universally collapse into a deterministic, near-perfect performance regime. As previewed in Figure~\ref{fig:illustration}, they fail to capture the true student spread, yielding JSD values orders of magnitude worse than the natural variance observed in the human held-out split (0.006).

\paragraph{The Paradox of Random Chance}
To contextualize this mismatch, we evaluate a naive random guessing baseline. Despite lacking cognitive capability and yielding a degraded mean accuracy (0.378), its wide, stochastic score distribution (red curve, Figure~\ref{fig:illustration}) paradoxically achieves a JSD (0.518) significantly closer to the human ground truth than any flagship LLM (0.634--0.773). That mindless guessing is distributionally closer to humans than state-of-the-art models highlights the core problem: LLM simulators suffer not from incapability, but from the lack of explicitly modeled cognitive limitations.

\paragraph{The Penalty of Model Capability}
Crucially, increasing LLM capability actually \emph{degrades} simulation fidelity. Across both open-weights (Llama, Mixtral) and proprietary models, scaling from efficient (GPT-4o-mini, Gemini-3.5-Flash-Lite) to frontier architectures (GPT-5.4, Gemini-3.7-Flash) pushes mean accuracy toward 0.98. This drives further divergence from the human baseline (JSD increases). Thus, standard persona prompting fails to restrict the model's global knowledge; the simulators trivially solve inferential and vocabulary tasks that demonstrably challenge young learners.

\begin{figure}[t]
    \centering
    \fbox{
        \begin{minipage}{0.95\linewidth}
            \scriptsize
            \ttfamily
            You are simulating a 5th-grade student in Norway taking a  reading comprehension test. Read the following text and answer the question as a typical 10-11 year old student would. The student may not always answer correctly.\\[1em]
            <TEXT>
            \textcolor{blue}{\{text\_content\}}
            </TEXT>\\[1em]
            <QUESTION>
            \textcolor{blue}{\{question\_text\}}
            </QUESTION>
        \end{minipage}
    }
    \caption{The baseline persona prompt template used in the pilot study. The exact formatting of \texttt{\{question\_text\}} for different question types (multiple choice, true-or-false, checkboxes) is in Appendix~\ref{app:question_prompts}.}
    \label{fig:baseline_prompt}
\end{figure}

\section{A Cognitively Bounded User Simulator}
\label{sec:cognitve_simulator}

As demonstrated in the pilot study, standard persona prompting systematically fails to capture the limitations of a developing reader, resulting in a ``superhuman bias'' where the LLM trivially performs global text synthesis. To overcome this, we propose shifting from superficial role-play to explicit cognitive modeling.
We conceptually frame our simulator as a \emph{cognitive language agent}~\citep{Sumers:2024:TMER}. In this architecture, the LLM is stripped of its unbounded access to the input context. Instead, it acts solely as the central executive (or processing hub), and its reasoning is strictly mediated by parameterized memory modules that enforce human-like cognitive bottlenecks~\citep{Baddeley:2000:TICS}.

\subsection{Grounding in Working Memory Capacity}

Our architecture is inspired by \citeauthor{Baddeley:2000:TICS}'s multicomponent model of working memory~\citep{Baddeley:2000:TICS}, and specifically by two of its components: the \emph{episodic buffer}, a limited-capacity store that integrates disparate pieces of information into a coherent representation, and the \emph{central executive}, which operates on that content to reason and answer. During reading comprehension, a student relies on the buffer to hold earlier text propositions available while processing new sentences---a prerequisite for resolving anaphora and drawing cross-sentence inferences (e.g., answering \emph{Interpret and Connect} questions). It is well established that this working-memory capacity constrains reading comprehension, and that the constraint is especially pronounced in developing readers, whose capacity is still maturing~\citep{DanemanCarpenter:1980:JVLVB,Cain:2004:JEP,Gathercole:2004:DP}.

Rather than model the buffer's detailed maintenance and integration functions, we deliberately \emph{simplify} it to a capacity-limited store, following \citeauthor{Cowan:2001:BBS}'s account in which working memory holds only a small number of chunks ($\sim$4 in adults)~\citep{Cowan:2001:BBS}. The critical distinction between an expert reader---or a standard LLM---and a young student is the strict capacity of this store: when a text requires integrating more propositions than it can hold, the reader experiences overload and falls back to a shallow, literal interpretation. The capacity limit introduced below operationalizes exactly this, with the LLM acting as the central executive that reasons over the bounded contents.

\subsection{Computational Implementation: The Episodic Bottleneck}

To computationally instantiate this cognitive bottleneck without requiring a complex, persistent multi-agent backend, we introduce a parameterized capacity limit, denoted as $C$. This parameter defines the maximum number of distinct text propositions the simulated student can actively maintain in working memory at any given time.

Furthermore, educational psychology suggests that developing readers employ varying test-taking strategies when approaching reading comprehension tasks~\citep{Ardoin:2024:LID}. To capture this behavioral variance, we operationalize the encoding process into two distinct variants:
\begin{enumerate}
    \item \textbf{Single-Pass Reading (SPR):} The simulator is provided only with the source text and extracts the $C$ most salient propositions overall. This simulates a student who reads the text once and attempts to answer all subsequent questions from this single, fixed memory trace.
    \item \textbf{Targeted Scanning (TS):} The simulator is provided with both the source text and a specific comprehension question. It extracts a maximum of $C$ distinct propositions that it deems most relevant to that specific question. This simulates a student employing a ``scan-and-search'' strategy for each item.
\end{enumerate}
Regardless of the chosen strategy, we operationalize this capacity limit by intercepting the LLM's standard generation process and forcing it through a constrained, two-stage reasoning pipeline (illustrated in Figure~\ref{fig:cbus_architecture}):

\begin{figure}[t]
    \centering
    \includegraphics[width=\linewidth]{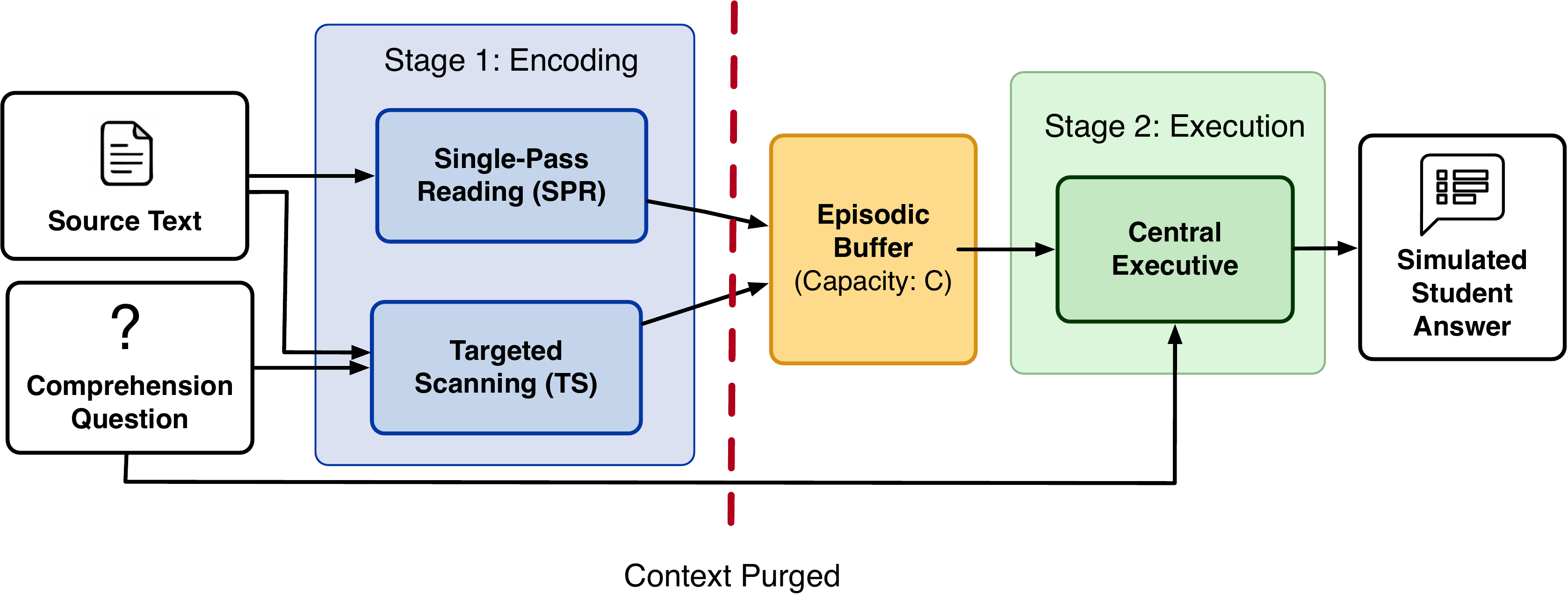}
    \caption{The Cognitively Bounded User Simulator (CBUS) architecture. In Stage 1 (Encoding), the simulator extracts a bounded number of propositions (Capacity: $C$) using either Single-Pass Reading (SPR) or Targeted Scanning (TS). The original source text is then purged from the context. In Stage 2 (Execution), the Central Executive answers the comprehension question relying solely on the restricted contents of the episodic buffer.}
    \label{fig:cbus_architecture}
\end{figure}

\begin{itemize}
    \item \textbf{Stage 1: Encoding (Filling the Buffer).} The LLM acts as the selective attention mechanism, executing either the SPR or TS strategy. During this stage, the model is explicitly forbidden from generating answers; it merely fills the buffer with $C$ extracted propositions.
    \item \textbf{Stage 2: Execution (Bounded Integration).} The original source text is completely purged from the LLM's context window. The central executive is then prompted (see Figure~\ref{fig:prompt_cbus_exec}) to answer the comprehension question utilizing \emph{only} the $C$ propositions successfully transferred into its simulated episodic buffer.
\end{itemize}
By structurally separating extraction from reasoning, we strictly bound the simulator's global attention mechanism. For example, if an inference question requires synthesizing three distinct facts from the text, but the simulator is parameterized with a buffer capacity of $C=2$, the agent is architecturally forced to attempt an answer using incomplete information. Consequently, the simulator's failures are no longer stochastic LLM hallucinations, but the direct result of verifiable cognitive constraints.

We fix $C$ from an independent cognitive principle rather than tuning it on human data. Following \citeauthor{Cowan:2001:BBS}'s embedded-processes model~\citep{Cowan:2001:BBS}, the capacity-limited store holds roughly four chunks in adults, and developmental estimates place a 10--11 year-old at or below this bound~\citep{Cowan:2016:EPT}; we therefore set $C=4$ for the holistic Single-Pass Reading strategy. Targeted Scanning instead models the narrow \emph{focus of attention} engaged during question-guided retrieval, which holds only a small number of items~\citep{Oberauer:2002:JEPLMC,Cowan:2001:BBS}; we set $C=2$---the smallest capacity that still permits integrating information from two points in a text, as \emph{Interpret and Connect} questions require. We report a full capacity sweep in Appendix~\ref{app:capacity}.


\begin{figure}[t]
    \centering
    \fbox{
        \begin{minipage}{0.95\linewidth}
            \scriptsize
            \ttfamily
            You are simulating a 5th-grade student in Norway taking a
            reading comprehension test. You are about to read a text,
            and you'll be asked questions about this text. As a young
            reader, you cannot remember everything from the text —
            only a few key facts.\\[1em]
            Read the text and identify at most \textcolor{blue}{\{capacity\}} distinct
            propositions (paraphrased atomic facts from the text) that
            you think are most relevant to the questions you'll be asked.
            Do not answer any questions yet.\\[1em]
            <TEXT>
            \textcolor{blue}{\{text\_content\}}
            </TEXT>\\[1em]
            Respond with ONLY a JSON object in the following format,
            with no other text:
            \{"propositions": ["proposition 1", "proposition 2", ...]\}
        \end{minipage}
    }
    \caption{Stage 1 Encoding Prompt for the \textbf{Single-Pass Reading (SPR)} strategy.}
    \label{fig:prompt_spr}
\end{figure}

\begin{figure}[t]
    \centering
    \fbox{
        \begin{minipage}{0.95\linewidth}
            \scriptsize
            \ttfamily
            You are simulating a 5th-grade student in Norway taking a
            reading comprehension test. You have just read the question
            below and are now scanning the text specifically to find the answer.
            As a young reader, you cannot process the whole text at once —
            you only focus on a few key details that seem to match the question.\\[1em]
            Scan the text and extract at most \textcolor{blue}{\{capacity\}} distinct propositions
            (paraphrased atomic facts from the text) that share keywords or
            seem most directly related to what the question is asking.
            Do NOT answer the question yet.\\[1em]
            <TEXT>
            \textcolor{blue}{\{text\_content\}}
            </TEXT>\\[1em]
            <QUESTION>
            \textcolor{blue}{\{question\_text\}}
            </QUESTION>\\[1em]
            Respond with ONLY a JSON object in the following format,
            with no other text:
            \{"propositions": ["proposition 1", "proposition 2", ...]\}
        \end{minipage}
    }
    \caption{Stage 1 Encoding Prompt for the \textbf{Targeted Scanning (TS)} strategy.}
    \label{fig:prompt_ts}
\end{figure}

\begin{figure}[t]
    \centering
    \fbox{
        \begin{minipage}{0.95\linewidth}
            \scriptsize
            \ttfamily
            You are simulating a 5th-grade student in Norway taking a
            reading comprehension test. You no longer have access to
            the full text — you can only rely on the propositions you
            remembered from reading. Answer the question as a typical
            10-11 year old student would, based ONLY on the
            propositions below. The student may not always answer
            correctly.\\[1em]
            <PROPOSITIONS>
            \textcolor{blue}{\{propositions\}}
            </PROPOSITIONS>\\[1em]
            <QUESTION>
            \textcolor{blue}{\{question\_text\}}
            </QUESTION>
        \end{minipage}
    }
    \caption{Stage 2 \textbf{Execution} Prompt for CBUS. The original text is withheld, forcing reliance on the bounded episodic buffer.}
    \label{fig:prompt_cbus_exec}
\end{figure}
\section{Experimental Setup}
\label{sec:exp_setup}

To isolate the impact of explicit cognitive modeling from underlying model capabilities, we evaluate our proposed Cognitively Bounded User Simulator (CBUS) framework against the standard zero-shot baseline persona simulator (detailed in Section~\ref{sec:baseline}) across a diverse suite of backbone LLMs.

\subsection{Evaluation Metrics}
\label{sec:exp_setup:metrics}

To rigorously evaluate simulation quality, we organize our metrics into two high-level categories: \emph{student-centric metrics}, which assess how well simulators approximate the overall performance and variance of the student population, and \emph{item-centric metrics}, which evaluate alignment on specific question difficulties. The formal definitions of the metrics is given in Appendix~\ref{app:metrics}.

\subsubsection{Student-Centric Metrics}
We employ three metrics to measure population-level fidelity. First, the \textbf{Absolute Performance Gap} calculates the absolute difference in overall mean accuracy between the human and simulator populations. Second, \textbf{Distributional Alignment} utilizes the Jensen-Shannon Divergence (JSD) to measure how accurately the simulator reflects the macro-level variance and shape of empirical student score distributions. Third, the \textbf{Expected Calibration Error (ECE)}~\citep{seshadri-etal-2026-lost} evaluates the alignment of success rates across discrete difficulty bins to capture non-uniform miscalibration (e.g., if a model overestimates easy questions but underestimates hard ones).

\subsubsection{Item-Centric Metrics}
To determine if simulators struggle with the same specific questions as real students, we calculate the empirical difficulty of each question (defined as its mean success rate across all subjects). We evaluate \textbf{Item-Level Difficulty Correlation} by computing both the Pearson ($\rho$) and Spearman's rank ($r_s$) correlation coefficients between the human and simulator question difficulties.

\subsection{Backbone LLMs and Inference Protocol}
To ensure our findings generalize across different architectures and scales, we use six distinct backbone LLMs, including both open-weight and proprietary models (details in Appendix~\ref{app:model_details}).
All models are evaluated in a zero-shot setting to prevent the simulators from memorizing specific student responses in-context. We set the generation temperature to $T=0.7$ on a standard $[0,1]$ scale to emulate the natural cognitive variance expected within a human student population.
To account for generation variance, we repeat all experiments $n=3$ times for every simulator-model pair and report the resulting averages.

\paragraph{Simulated Population Construction}
Each simulated population is a one-to-one mirror of the real Ground Truth split. For every real student, we retain their exact set of questions (i.e., those associated with the self-selected texts) remove their answers, and task the simulator with answering those questions. Consequently, the simulated and empirical populations are matched on exposure by construction: both consist of 1{,}198 subjects who answer the same set of questions. All student-centric metrics compare the resulting distributions of per-subject scores between the simulated and real \emph{populations}, rather than pairing individual subjects.

\section{Results and Analysis}
\label{sec:results_and_analysis}

\begin{table*}[t]
    \caption{Evaluation results across different backbone LLMs and simulation variants, under independent per-student sampling (SPR $C{=}4$, TS $C{=}2$). Results are reported as reported as mean $\pm$ standard deviation (in smaller font) over $n=3$ runs. For each metric, the best simulated result is highlighted in \textbf{bold} and the second best in \emph{italics} (\emph{Real students} and \emph{Random Baseline} are references and are not ranked).}
    \label{tab:results}
    \centering
    \scriptsize
    \begin{tabular}{ll r@{\,}l r@{\,}l r@{\,}l r@{\,}l r@{\,}l}
        \toprule
                                                     &                  & \multicolumn{6}{c}{\textbf{Student-Centric}}         & \multicolumn{4}{c}{\textbf{Item-Centric}}                                                                                                                                                                                                                                                              \\
        \cmidrule(lr){3-8} \cmidrule(lr){9-12}
        \textbf{Backbone}                            & \textbf{Variant} & \multicolumn{2}{c}{\textbf{Abs. Gap} ($\downarrow$)} & \multicolumn{2}{c}{\textbf{JSD} ($\downarrow$)} & \multicolumn{2}{c}{\textbf{ECE} ($\downarrow$)} & \multicolumn{2}{c}{\textbf{Pearson $\rho$} ($\uparrow$)} & \multicolumn{2}{c}{\textbf{Spearman $r_s$} ($\uparrow$)}                                                                                \\
        \midrule
        \multicolumn{2}{l}{Real students (Held-out)} & 0.026            &                                                      & 0.006                                           &                                                 & 0.027                                                    &                                                          & 0.895        &                & 0.845        &                               \\
        \midrule
        \multicolumn{2}{l}{Random Baseline}          & 0.309            & \stdev{.001}                                         & 0.518                                           & \stdev{.004}                                    & 0.306                                                    & \stdev{.001}                                             & 0.409        & \stdev{.014}   & 0.319        & \stdev{.008}                  \\
        \midrule
        Llama-3.3-70b-Instruct                       & baseline         & 0.279                                                & \stdev{.001}                                    & 0.679                                           & \stdev{.011}                                             & 0.285                                                    & \stdev{.001} & 0.241          & \stdev{.003} & 0.179          & \stdev{.045} \\
                                                     & CBUS-TS          & 0.238                                                & \stdev{.001}                                    & 0.475                                           & \stdev{.007}                                             & 0.244                                                    & \stdev{.001} & \textbf{0.455} & \stdev{.003} & 0.304          & \stdev{.016} \\
                                                     & CBUS-SPR         & \emph{0.037}                                         & \stdev{.002}                                    & \emph{0.093}                                    & \stdev{.007}                                             & \textbf{0.069}                                           & \stdev{.002} & 0.301          & \stdev{.007} & 0.297          & \stdev{.004} \\
        \midrule
        Mixtral-8x22b-Instruct                       & baseline         & 0.269                                                & \stdev{.000}                                    & 0.634                                           & \stdev{.003}                                             & 0.276                                                    & \stdev{.000} & 0.326          & \stdev{.002} & 0.289          & \stdev{.020} \\
                                                     & CBUS-TS          & 0.237                                                & \stdev{.001}                                    & 0.467                                           & \stdev{.001}                                             & 0.243                                                    & \stdev{.001} & 0.417          & \stdev{.009} & 0.314          & \stdev{.020} \\
                                                     & CBUS-SPR         & 0.124                                                & \stdev{.002}                                    & 0.170                                           & \stdev{.006}                                             & 0.128                                                    & \stdev{.002} & 0.377          & \stdev{.004} & \textbf{0.370} & \stdev{.016} \\
        \midrule
        GPT-4o-mini                                  & baseline         & 0.271                                                & \stdev{.001}                                    & 0.646                                           & \stdev{.002}                                             & 0.278                                                    & \stdev{.000} & 0.315          & \stdev{.002} & 0.240          & \stdev{.008} \\
                                                     & CBUS-TS          & 0.226                                                & \stdev{.001}                                    & 0.418                                           & \stdev{.005}                                             & 0.233                                                    & \stdev{.000} & 0.415          & \stdev{.003} & 0.301          & \stdev{.019} \\
                                                     & CBUS-SPR         & 0.096                                                & \stdev{.001}                                    & 0.122                                           & \stdev{.004}                                             & 0.100                                                    & \stdev{.001} & 0.355          & \stdev{.004} & \emph{0.328}   & \stdev{.004} \\
        \midrule
        GPT-5.4                                      & baseline         & 0.287                                                & \stdev{.000}                                    & 0.732                                           & \stdev{.001}                                             & 0.293                                                    & \stdev{.000} & 0.255          & \stdev{.001} & 0.172          & \stdev{.002} \\
                                                     & CBUS-TS          & 0.271                                                & \stdev{.001}                                    & 0.618                                           & \stdev{.001}                                             & 0.276                                                    & \stdev{.001} & 0.397          & \stdev{.004} & 0.286          & \stdev{.005} \\
                                                     & CBUS-SPR         & 0.129                                                & \stdev{.000}                                    & 0.188                                           & \stdev{.001}                                             & 0.131                                                    & \stdev{.001} & 0.224          & \stdev{.006} & 0.225          & \stdev{.014} \\
        \midrule
        Gemini-3.5-Flash-Lite                        & baseline         & 0.285                                                & \stdev{.000}                                    & 0.713                                           & \stdev{.004}                                             & 0.291                                                    & \stdev{.000} & 0.290          & \stdev{.001} & 0.220          & \stdev{.008} \\
                                                     & CBUS-TS          & 0.238                                                & \stdev{.001}                                    & 0.470                                           & \stdev{.007}                                             & 0.243                                                    & \stdev{.001} & \emph{0.454}   & \stdev{.002} & 0.298          & \stdev{.009} \\
                                                     & CBUS-SPR         & \textbf{0.031}                                       & \stdev{.002}                                    & \textbf{0.086}                                  & \stdev{.004}                                             & \emph{0.074}                                             & \stdev{.001} & 0.304          & \stdev{.001} & 0.295          & \stdev{.007} \\
        \midrule
        Gemini-3.7-Flash                             & baseline         & 0.298                                                & \stdev{.000}                                    & 0.773                                           & \stdev{.001}                                             & 0.305                                                    & \stdev{.000} & 0.246          & \stdev{.001} & 0.154          & \stdev{.006} \\
                                                     & CBUS-TS          & 0.281                                                & \stdev{.000}                                    & 0.668                                           & \stdev{.004}                                             & 0.287                                                    & \stdev{.000} & 0.285          & \stdev{.001} & 0.195          & \stdev{.011} \\
                                                     & CBUS-SPR         & 0.091                                                & \stdev{.000}                                    & 0.120                                           & \stdev{.003}                                             & 0.115                                                    & \stdev{.000} & 0.236          & \stdev{.002} & 0.202          & \stdev{.002} \\
        \bottomrule
    \end{tabular}
\end{table*}

Table~\ref{tab:results} presents the evaluation of our cognitive bottleneck framework (CBUS) against standard persona-prompted baselines. As previously described, we evaluate two behavioral variants: Single-Pass Reading (SPR) and Targeted Scanning (TS). Rather than tuning the capacity on human response data, we fix $C$ from an independent cognitive principle (Section~\ref{sec:cognitve_simulator}): $C=4$ for SPR and $C=2$ for TS. A sensitivity analysis over a broad range of capacities (Appendix~\ref{app:capacity}) confirms that these principled values lie at or near the empirical optimum.

\paragraph{Closing the Simulation Gap}
The primary finding is that explicitly modeling cognitive boundaries successfully mitigates the ``superhuman bias'' of LLMs. Across all evaluated backbones and test-taking strategies, the CBUS framework consistently improves upon the baseline across all student-centric metrics. By restricting the episodic buffer, we successfully bring the simulated score distributions much closer to real student behavior.

\paragraph{Trade-offs in Test-Taking Strategies}
While CBUS improves global simulation fidelity, a clear trade-off emerges between the two encoding strategies. SPR drastically outperforms TS on all student-centric metrics, producing the most realistic macro-level distributions. TS, in turn, yields the strongest item-centric alignment: it improves both Pearson and Spearman correlations over the baseline for every backbone and attains the highest Pearson correlation of any variant across all backbones, indicating that it better captures the specific question difficulties experienced by real students. Notably, SPR also improves item-centric correlations over the baseline in most cases, so the two strategies are complementary rather than strictly opposed.

\paragraph{Model Capability vs. Simulation Fidelity}
Interestingly, utilizing more capable, state-of-the-art proprietary LLMs (e.g., GPT-5.4, Gemini-3.7-Flash) does not organically translate to more realistic student simulations; the strongest results instead come from smaller or open-weights backbones. Under SPR, Gemini-3.5-Flash-Lite and Llama-3.3-70B produce the closest student-centric alignment (with an absolute gap of $0.031$ and $0.037$, respectively), while the best item-centric correlations are obtained under TS, led by Llama-3.3-70B and Gemini-3.5-Flash-Lite on Pearson and Mixtral-8x22B on Spearman. This suggests that alignment for human simulation depends more on the architectural constraints of the simulator than on the raw reasoning power of the underlying backbone. This corroborates \citet{Zhou:2026:COLM}, who report that ``higher general model capability does not necessarily yield more faithful user simulation.''

\paragraph{Diagnostic Baselines and Ablation}
To isolate \emph{why} CBUS works, we compare it against three diagnostic baselines on two backbones (Llama-3.3-70B and Gemini-3.5-Flash-Lite; full results in Appendix~\ref{app:diagnostics}). The first two try to lower the simulator's accuracy to the empirical student performance level without any cognitive mechanism: a \emph{Calibrated Low-persona} prompt instructing the model to act as a below-average student, and \emph{Answer-noising}, which randomly flips correct baseline answers until the mean matches the student population. Neither reproduces the students' score distributions: answer-noising matches the mean by construction yet leaves the item-difficulty correlation near zero, since uniform flipping carries no information about which questions are hard, while the low-persona prompt is unreliable---one LLM backbone largely ignores it and stays superhuman, and where it does comply it still falls short of CBUS. This confirms that matching aggregate accuracy is neither reliably achievable by prompting nor sufficient for distributional or item-level fidelity. Finally, a \emph{Random-dropout} ablation replaces CBUS-SPR's salience-based proposition selection with random dropout at the same capacity; it performs on par with CBUS-SPR (better on some metrics, worse on others), indicating that the capacity \emph{bound} itself, rather than the specific salience heuristic used to fill it, is the primary driver of the improvement.

\paragraph{Impact Across Question Categories}
We further break down performance by question category---\emph{Locate} versus \emph{Interpret \& Reflect} (combining ``Interpret \& Connect'' and ``Reflect \& Evaluate'')---for Llama-3.3-70B and Gemini-3.5-Flash-Lite (Appendix~\ref{app:results}). The superhuman bias is most pronounced on retrieval-oriented \emph{Locate} questions, which the baselines answer almost uniformly correctly and therefore fail to capture the natural spread of student scores. The best CBUS strategy then depends on the task: SPR dominates the student-centric metrics on \emph{Locate}---for Llama it cuts the JSD score from $0.640$ to $0.078$. TS  best captures item-level difficulty on higher-order \emph{Interpret \& Reflect} questions, where Llama's item-centric Pearson correlation rises to $\rho = 0.556$ from the baseline $0.279$.

\section{Conclusion}
\label{sec:concl}
In this work, we investigated LLM-based user simulators in the setting of simulating the reading comprehension skills of primary-school students, demonstrating that standard persona prompting results in a severe ``superhuman bias'' that fails to capture the natural variance of developing readers.
To bridge this gap, we introduced CBUS, a framework grounded in cognitive psychology that forces the LLM to operate under explicit, limited-capacity memory bottlenecks. Within this framework, we modeled two distinct test-taking strategies. By evaluating this approach against a massive dataset of real student responses, we demonstrated that explicitly modeling cognitive boundaries successfully narrows the simulation gap across a diverse set of LLM backbones, yielding improvements in both student-centric and item-centric metrics.
Notably, our modeling of different test-taking strategies revealed distinct alignment trade-offs that warrant further investigation in future work. Furthermore, our highest simulation fidelity was achieved with smaller, less capable models rather than the most capable frontier LLMs, suggesting that building realistic user simulators relies far more on explicitly modeling the architectural constraints of human cognition than on scaling raw model intelligence.

\section*{Limitations}

While the CBUS framework successfully demonstrates that explicitly modeling cognitive bounds improves simulation fidelity, our approach has several limitations.

First, our framework focuses exclusively on a single cognitive dimension: the capacity bottleneck of working memory. Real-world student performance is heavily influenced by complex non-cognitive factors, including test anxiety, fatigue, and intrinsic motivation. Although our dataset collection protocol attempted to foster engagement by allowing students to choose between two randomly paired texts, this binary choice does not guarantee genuine interest. Consequently, our current simulator lacks the mechanisms to model performance degradation caused by attention drift, apathy, or physical fatigue.

Relatedly, CBUS bounds the \emph{amount} of information a reader retains, but idealizes \emph{which} information is kept and \emph{how faithfully}. The encoding stage uses a capable LLM to select the most salient (SPR) or question-relevant (TS) propositions and store them as accurate paraphrases; it therefore caps capacity without reproducing the selection and encoding errors of a real 10--11 year-old, who may retain irrelevant details, miss the gist, or misremember. Likewise, the execution stage still draws on the backbone's general language knowledge and answer-elimination strategies. Our evidence thus supports the narrower claim that restricting the quantity of accessible information improves distributional and item-level alignment with real students, rather than the stronger claim that CBUS faithfully reproduces the process by which a child reads and remembers. Modeling encoding noise and imperfect, non-expert fact selection is a natural next step.

Second, our implementation models Single-Pass Reading (SPR) and Targeted Scanning (TS) as rigid, mutually exclusive strategies. In reality, developing readers often employ dynamic, hybrid approaches---perhaps reading the text holistically once, but reverting to targeted scanning for highly specific or challenging questions. Future iterations of CBUS should explore dynamic strategy-switching mechanisms and probabilistic capacity allocation to better reflect fluid human test-taking behavior.

Finally, our evaluation scope is constrained by task format and demographics. To ensure an objective, automatically gradable ground truth, we deliberately binarized responses and excluded free-text generation tasks. Therefore, our findings regarding episodic bottlenecks apply primarily to recognition and retrieval tasks rather than generative comprehension. Furthermore, while our dataset provides high-quality, standardized empirical data, it is restricted to 4th to 6th-grade Norwegian students. Validating the universality of these simulated cognitive bounds across different languages, educational systems, and developmental stages remains an important direction for future work.

\bibliography{arr2026-simreading}

@article{Ardoin:2024:LID,
  author  = {Scott P. Ardoin and Katherine S. Binder and Paulina A. Kulesz and Eloise Nimocks and Joshua A. Mellott},
  title   = {Examining the influence of passage and student characteristics on test-taking strategies: An eye-tracking study},
  journal = {Learning and Individual Differences},
  volume  = {109},
  pages   = {102386},
  year    = {2024},
  issn    = {1041-6080},
  doi     = {https://doi.org/10.1016/j.lindif.2023.102386},
  url     = {https://www.sciencedirect.com/science/article/pii/S1041608023001309}
}

@article{Argyle:2023:PA,
  title   = {Out of One, Many: Using Language Models to Simulate Human Samples},
  author  = {Argyle, Lisa P. and Busby, Ethan C. and Fulda, Nancy and Gubler, Joshua R. and Rytting, Christopher and Wingate, David},
  journal = {Political Analysis},
  year    = {2023},
  volume  = {31},
  number  = {3},
  pages   = {337--351}
}

@article{Baddeley:2000:TICS,
  title   = {The episodic buffer: a new component of working memory?},
  author  = {Baddeley, Alan},
  journal = {Trends in Cognitive Sciences},
  volume  = {4},
  number  = {11},
  pages   = {417--423},
  year    = {2000},
  doi     = {https://doi.org/10.1016/s1364-6613(00)01538-2}
}

@article{Balog:2024:FnTIR,
  author  = {Krisztian Balog and ChengXiang Zhai},
  title   = {User Simulation for Evaluating Information Access Systems},
  journal = {Foundations and Trends in Information Retrieval},
  year    = {2024},
  volume  = {18},
  doi     = {10.1561/1500000098},
  url     = {http://dx.doi.org/10.1561/1500000098},
  issn    = {1554-0669},
  number  = {1-2},
  pages   = {1-261}
}

@inproceedings{Balog:2025:SIGIRb,
  author    = {Balog, Krisztian and Bernard, Nolwenn and Zerhoudi, Saber and Zhai, ChengXiang},
  title     = {Theory and Toolkits for User Simulation in the Era of Generative AI: User Modeling, Synthetic Data Generation, and System Evaluation},
  year      = {2025},
  url       = {https://doi.org/10.1145/3726302.3731697},
  doi       = {10.1145/3726302.3731697},
  booktitle = {Proceedings of the 48th International ACM SIGIR Conference on Research and Development in Information Retrieval},
  pages     = {4138–4141},
  series    = {SIGIR '25}
}

@article{Cain:2004:JEP,
  title   = {Children's reading comprehension ability: Concurrent prediction by working memory, verbal ability, and component skills},
  author  = {Cain, Kate and Oakhill, Jane and Bryant, Peter},
  journal = {Journal of Educational Psychology},
  volume  = {96},
  number  = {1},
  pages   = {31--42},
  year    = {2004},
  doi     = {https://doi.org/10.1037/0022-0663.96.1.31}
}

@article{Cowan:2001:BBS,
  title   = {The magical number 4 in short-term memory: A reconsideration of mental storage capacity},
  author  = {Cowan, Nelson},
  journal = {Behavioral and Brain Sciences},
  volume  = {24},
  number  = {1},
  pages   = {87--114},
  year    = {2001},
  doi     = {https://doi.org/10.1017/s0140525x01003922}
}

@article{Cowan:2016:EPT,
  title   = {Working memory maturation: Can we get at the essence of cognitive growth?},
  author  = {Cowan, Nelson},
  journal = {Perspectives on Psychological Science},
  volume  = {11},
  number  = {2},
  pages   = {239--264},
  year    = {2016},
  doi     = {https://doi.org/10.1177/1745691615621279}
}

@article{DanemanCarpenter:1980:JVLVB,
  title   = {Individual differences in working memory and reading},
  author  = {Daneman, Meredyth and Carpenter, Patricia A.},
  journal = {Journal of Verbal Learning and Verbal Behavior},
  volume  = {19},
  number  = {4},
  pages   = {450--466},
  year    = {1980},
  doi     = {https://doi.org/10.1016/S0022-5371(80)90312-6}
}

@article{Davidson:2023:arXiv,
  title   = {User Simulation with Large Language Models for Evaluating Task-Oriented Dialogue},
  author  = {Sam Davidson and Salvatore Romeo and Raphael Shu and James Gung and Arshit Gupta and Saab Mansour and Yi Zhang},
  year    = {2023},
  volume  = {cs.CL/2309.13233},
  journal = {arXiv}
}

@inproceedings{Gao:2025:AAAI,
  title     = {{Agent4Edu}: Generating Learner Response Data by Generative Agents for Intelligent Education Systems},
  author    = {Gao, Weibo and Liu, Qi and others},
  booktitle = {Proceedings of the Thirty-Ninth AAAI Conference on Artificial Intelligence and Thirty-Seventh Conference on Innovative Applications of Artificial Intelligence and Fifteenth Symposium on Educational Advances in Artificial Intelligence},
  articleno = {2667},
  numpages  = {10},
  series    = {AAAI'25/IAAI'25/EAAI'25},
  year      = {2025},
  doi       = {10.1609/aaai.v39i22.34565}
}

@article{Gathercole:2004:DP,
  title   = {The structure of working memory from 4 to 15 years of age},
  author  = {Gathercole, Susan E. and Pickering, Susan J. and Ambridge, Benjamin and Wearing, Hannah},
  journal = {Developmental Psychology},
  volume  = {40},
  number  = {2},
  pages   = {177--190},
  year    = {2004},
  doi     = {https://doi.org/10.1037/0012-1649.40.2.177}
}

@inproceedings{Kiesel:2024:ECIR,
  title     = {Simulating Follow-up Questions in Conversational Search},
  author    = {Johannes Kiesel and Marcel Gohsen and Nailia Mirzakhmedova and Matthias Hagen and Benno Stein},
  booktitle = {46th European Conference on IR Research},
  year      = {2024}
}

@inproceedings{Naous:2026:ICLR,
  title     = {Flipping the Dialogue: Training and Evaluating User Language Models},
  author    = {Tarek Naous and Philippe Laban and Wei Xu and Jennifer Neville},
  booktitle = {The Fourteenth International Conference on Learning Representations},
  year      = {2026},
  series    = {ICLR '26},
  url       = {https://openreview.net/forum?id=ykSmkVqzn4}
}

@book{NESH2021,
  author    = {NESH, Den nasjonale forskningsetiske komité for samfunnsvitenskap og humaniora},
  title     = {Guidelines for Research Ethics in the Social Sciences and the Humanities},
  publisher = {De nasjonale forskningsetiske komiteene},
  year      = {2021},
  note      = {5th edition, updated 2023},
  url       = {https://www.forskningsetikk.no/retningslinjer/hum-sam/forskningsetiske-retningslinjer-for-samfunnsvitenskap-og-humaniora/},
  address   = {Oslo, Norway}
}

@article{Oberauer:2002:JEPLMC,
  title   = {Access to information in working memory: Exploring the focus of attention},
  author  = {Oberauer, Klaus},
  journal = {Journal of Experimental Psychology: Learning, Memory, and Cognition},
  volume  = {28},
  number  = {3},
  pages   = {411--421},
  year    = {2002},
  url     = {https://pubmed.ncbi.nlm.nih.gov/12018494/}
}

@inproceedings{Park:2023:UIST,
  author    = {Park, Joon Sung and O'Brien, Joseph and Cai, Carrie Jun and Morris, Meredith Ringel and Liang, Percy and Bernstein, Michael S.},
  title     = {Generative Agents: Interactive Simulacra of Human Behavior},
  year      = {2023},
  booktitle = {Proceedings of the 36th Annual ACM Symposium on User Interface Software and Technology},
  articleno = {2},
  series    = {UIST '23},
  doi       = {https://doi.org/10.1145/3586183.3606763}
}

@article{Piao:2026:iFuture,
  author  = {Jinghua Piao and Yuwei Yan and Jun Zhang and Nian Li and Junbo Yan and Xiaochong Lan and Zhihong Lu and Zhiheng Zheng and Jing Yi Wang and Di Zhou and Chen Gao and Fengli Xu and Fang Zhang and Ke Rong and Jun Su and Yong Li},
  title   = {AgentSociety: Large-scale simulation of LLM-driven generative agents advances understanding of human behaviors and society},
  year    = {2026},
  journal = {iFuture},
  url     = {https://www.sciopen.com/article/10.26599/IF.2026.9710004},
  doi     = {10.26599/IF.2026.9710004}
}

@article{Sabatini:2024:PIRLS,
  author    = {Sabatini, J. and Kennedy, A. and Wry, E. and von Davier, M.},
  editor    = {von Davier, M. and Kennedy, A.},
  title     = {{PIRLS} 2026 reading assessment framework},
  journal   = {{PIRLS} 2026 Assessment Frameworks},
  publisher = {Boston College, TIMSS \& PIRLS International Study Center},
  doi       = {10.6017/lse.tpisc.tr2103.kb5202},
  year      = {2024}
}

@inproceedings{Shim:2026:ICLR,
  title     = {Non-Collaborative User Simulators for Tool Agents},
  author    = {Jeonghoon Shim and Woojung Song and Cheyon Jin and Seungwon Kook and Yohan Jo},
  booktitle = {The Fourteenth International Conference on Learning Representations},
  year      = {2026},
  url       = {https://openreview.net/forum?id=UAUimofy3W}
}

@misc{Sonkar:2024:arXiv,
  title         = {{LLM}-based Cognitive Models of Students with Misconceptions},
  author        = {Sonkar, Shashank and Chen, Xinghe and Liu, Naiming and Baraniuk, Richard G. and Sachan, Mrinmaya},
  year          = {2024},
  eprint        = {2410.12294},
  archiveprefix = {arXiv},
  note          = {arXiv:2410.12294}
}

@article{Sumers:2024:TMER,
  title   = {Cognitive Architectures for Language Agents},
  author  = {Theodore Sumers and Shunyu Yao and Karthik R Narasimhan and Thomas L. Griffiths},
  journal = {Transactions on Machine Learning Research},
  year    = {2024},
  url     = {https://openreview.net/forum?id=1i6ZCvflQJ}
}

@misc{Terragni:2023:arXiv,
  title         = {In-Context Learning User Simulators for Task-Oriented Dialog Systems},
  author        = {Silvia Terragni and Modestas Filipavicius and Nghia Khau and Bruna Guedes and André Manso and Roland Mathis},
  year          = {2023},
  eprint        = {2306.00774},
  archiveprefix = {arXiv},
  primaryclass  = {cs.CL},
  url           = {https://arxiv.org/abs/2306.00774}
}

@misc{udir_framework_reading,
  author       = {{Utdanningsdirektoratet}},
  title        = {Rammeverk for grunnleggende ferdigheter: 2.3 {\AA} kunne lese},
  organization = {Utdanningsdirektoratet},
  url          = {https://www.udir.no/laring-og-trivsel/rammeverk/rammeverk-for-grunnleggende-ferdigheter/2.3-a-kunne-lese},
  year         = 2017
}

@incollection{vonDavier:2024:Bookchapter,
  title     = {How will {AI} change adaptive testing?},
  author    = {von Davier, Alina A},
  booktitle = {Research for Practical Issues and Solutions in Computerized Multistage Testing},
  pages     = {460--480},
  year      = {2024},
  publisher = {Routledge}
}

@incollection{Wainer:2000:Bookchapter,
  title     = {Item Response Theory, Item Calibration, and Proficiency Estimation},
  author    = {Wainer, Howard and Mislevy, Robert J},
  booktitle = {Computerized adaptive testing},
  pages     = {61--100},
  year      = {2000},
  publisher = {Routledge}
}

@inproceedings{Wu:2026:ICML,
  title     = {Human{LM}: Simulating Users with State Alignment Beats Response Imitation},
  author    = {Shirley Wu and Evelyn Choi and Arpandeep Khatua and Zhanghan Wang and Joy He-Yueya and Tharindu Cyril Weerasooriya and Wei Wei and Diyi Yang and Jure Leskovec and James Zou},
  booktitle = {Forty-third International Conference on Machine Learning},
  year      = {2026},
  url       = {https://openreview.net/forum?id=xyFVjBi9wC}
}

@inproceedings{Wang:2024:WWW,
  author    = {Wang, Zhenduo and Xu, Zhichao and Srikumar, Vivek and Ai, Qingyao},
  title     = {An In-depth Investigation of User Response Simulation for Conversational Search},
  year      = {2024},
  booktitle = {ACM on Web Conference 2024}
}

@inproceedings{Xiao:2024:LAS,
  author    = {Lu, Xinyi and Wang, Xu},
  title     = {Generative Students: Using LLM-Simulated Student Profiles to Support Question Item Evaluation},
  year      = {2024},
  doi       = {10.1145/3657604.3662031},
  booktitle = {Proceedings of the Eleventh ACM Conference on Learning @ Scale},
  pages     = {16–27},
  numpages  = {12},
  series    = {L@S '24}
}

@misc{Xu:2024:arXiv,
  title         = {EduAgent: Generative Student Agents in Learning},
  author        = {Xu, Songlin and Zhang, Xinyu and Qin, Lianhui},
  year          = {2024},
  eprint        = {2404.07963},
  archiveprefix = {arXiv},
  note          = {arXiv:2404.07963}
}

@inproceedings{Zerhoudi:2024:JCDL,
  author    = {Zerhoudi, Saber and Granitzer, Michael},
  title     = {Cognitive-Aware User Search Behavior Simulation},
  year      = {2025},
  url       = {https://doi.org/10.1145/3677389.3702598},
  doi       = {10.1145/3677389.3702598},
  booktitle = {Proceedings of the 24th ACM/IEEE Joint Conference on Digital Libraries},
  articleno = {20},
  series    = {JCDL '24}
}

@inproceedings{Zhang:2025:SIGIR,
  author    = {Zhang, Erhan and Wang, Xingzhu and Gong, Peiyuan and Yang, Zixuan and Mao, Jiaxin},
  title     = {Exploring Human-Like Thinking in Search Simulations with Large Language Models},
  year      = {2025},
  url       = {https://doi.org/10.1145/3726302.3730193},
  doi       = {10.1145/3726302.3730193},
  booktitle = {Proceedings of the 48th International ACM SIGIR Conference on Research and Development in Information Retrieval},
  pages     = {2669–2673},
  series    = {SIGIR '25}
}

@inproceedings{Zhou:2026:COLM,
  title     = {Mind the Sim2Real Gap in User Simulation for Agentic Tasks},
  author    = {Xuhui Zhou and Weiwei Sun and Qianou Ma and Yiqing Xie and Jiarui Liu and Weihua Du and Sean Welleck and Yiming Yang and Graham Neubig and Sherry Tongshuang Wu and Maarten Sap},
  booktitle = {The Third Annual Conference on Language Modeling},
  year      = {2026},
  series    = {COLM '26},
  url       = {https://arxiv.org/abs/2603.11245}
}

@inproceedings{ferreira-etal-2024-multi,
  title     = {Multi-trait User Simulation with Adaptive Decoding for Conversational Task Assistants},
  author    = {Ferreira, Rafael and Semedo, David and Magalhaes, Joao},
  editor    = {Al-Onaizan, Yaser and Bansal, Mohit and Chen, Yun-Nung},
  booktitle = {Findings of the Association for Computational Linguistics: EMNLP 2024},
  month     = nov,
  year      = {2024},
  address   = {Miami, Florida, USA},
  publisher = {Association for Computational Linguistics},
  url       = {https://aclanthology.org/2024.findings-emnlp.945/},
  doi       = {10.18653/v1/2024.findings-emnlp.945},
  pages     = {16105--16130}
}

@inproceedings{kolluri-etal-2025-finetuning,
  title     = {Finetuning {LLM}s for Human Behavior Prediction in Social Science Experiments},
  author    = {Kolluri, Akaash  and
               Wu, Shengguang  and
               Park, Joon Sung  and
               Bernstein, Michael S.},
  editor    = {Christodoulopoulos, Christos  and
               Chakraborty, Tanmoy  and
               Rose, Carolyn  and
               Peng, Violet},
  booktitle = {Proceedings of the 2025 Conference on Empirical Methods in Natural Language Processing},
  month     = nov,
  year      = {2025},
  address   = {Suzhou, China},
  publisher = {Association for Computational Linguistics},
  url       = {https://aclanthology.org/2025.emnlp-main.1530/},
  doi       = {10.18653/v1/2025.emnlp-main.1530},
  pages     = {30096--30111}
}

@inproceedings{meshi-etal-2026-convapparel,
  title     = {{C}onv{A}pparel: A Benchmark Dataset and Validation Framework for User Simulators in Conversational Recommenders},
  author    = {Meshi, Ofer and Balog, Krisztian and Goldman, Sally and Caciularu, Avi and Tennenholtz, Guy and Jeong, Jihwan and Globerson, Amir and Boutilier, Craig},
  editor    = {Demberg, Vera and Inui, Kentaro and Marquez, Llu{\'i}s},
  booktitle = {Proceedings of the 19th Conference of the {E}uropean Chapter of the {A}ssociation for {C}omputational {L}inguistics (Volume 1: Long Papers)},
  month     = mar,
  year      = {2026},
  address   = {Rabat, Morocco},
  publisher = {Association for Computational Linguistics},
  url       = {https://aclanthology.org/2026.eacl-long.244/},
  doi       = {10.18653/v1/2026.eacl-long.244},
  pages     = {5270--5304},
  isbn      = {979-8-89176-380-7}
}

@inproceedings{seshadri-etal-2026-lost,
  title     = {Lost in Simulation: {LLM}-Simulated Users are Unreliable Proxies for Human Users in Agentic Evaluations},
  author    = {Seshadri, Preethi  and
               Cahyawijaya, Samuel  and
               Odumakinde, Ayomide  and
               Singh, Sameer  and
               Goldfarb-Tarrant, Seraphina},
  editor    = {Liakata, Maria  and
               Moreira, Viviane P.  and
               Zhang, Jiajun  and
               Jurgens, David},
  booktitle = {Proceedings of the 64th Annual Meeting of the {A}ssociation for {C}omputational {L}inguistics (Volume 1: Long Papers)},
  month     = jul,
  year      = {2026},
  address   = {San Diego, California, United States},
  publisher = {Association for Computational Linguistics},
  url       = {https://aclanthology.org/2026.acl-long.2192/},
  doi       = {10.18653/v1/2026.acl-long.2192},
  pages     = {47423--47439}
}

@inproceedings{wang-etal-2025-know,
  title     = {Know You First and Be You Better: Modeling Human-Like User Simulators via Implicit Profiles},
  author    = {Wang, Kuang and Li, Xianfei and Yang, Shenghao and Zhou, Li and Jiang, Feng and Li, Haizhou},
  editor    = {Che, Wanxiang and Nabende, Joyce and Shutova, Ekaterina and Pilehvar, Mohammad Taher},
  booktitle = {Proceedings of the 63rd Annual Meeting of the Association for Computational Linguistics (Volume 1: Long Papers)},
  month     = jul,
  year      = {2025},
  address   = {Vienna, Austria},
  publisher = {Association for Computational Linguistics},
  url       = {https://aclanthology.org/2025.acl-long.1025/},
  doi       = {10.18653/v1/2025.acl-long.1025},
  pages     = {21082--21107},
  isbn      = {979-8-89176-251-0}
}

@inproceedings{yancey-etal-2024-bert,
  title     = {{BERT}-{IRT}: Accelerating Item Piloting with {BERT} Embeddings and Explainable {IRT} Models},
  author    = {Yancey, Kevin P.  and
               Runge, Andrew  and
               LaFlair, Geoffrey  and
               Mulcaire, Phoebe},
  editor    = {Kochmar, Ekaterina  and
               Bexte, Marie  and
               Burstein, Jill  and
               Horbach, Andrea  and
               Laarmann-Quante, Ronja  and
               Tack, Ana{\"i}s  and
               Yaneva, Victoria  and
               Yuan, Zheng},
  booktitle = {Proceedings of the 19th Workshop on Innovative Use of NLP for Building Educational Applications (BEA 2024)},
  month     = jun,
  year      = {2024},
  address   = {Mexico City, Mexico},
  publisher = {Association for Computational Linguistics},
  url       = {https://aclanthology.org/2024.bea-1.35/},
  pages     = {428--438}
}

@inproceedings{yang-etal-2025-consistent,
  title     = {Consistent Client Simulation for Motivational Interviewing-based Counseling},
  author    = {Yang, Yizhe and Achananuparp, Palakorn and Huang, Heyan and Jiang, Jing and Lim, Nicholas Gabriel and Ern, Cameron Tan Shi and Kit, Phey Ling and Xiuhui, Jenny Giam and Pinto, John and Lim, Ee-Peng},
  editor    = {Che, Wanxiang and Nabende, Joyce and Shutova, Ekaterina and Pilehvar, Mohammad Taher},
  booktitle = {Proceedings of the 63rd Annual Meeting of the Association for Computational Linguistics (Volume 1: Long Papers)},
  month     = jul,
  year      = {2025},
  address   = {Vienna, Austria},
  publisher = {Association for Computational Linguistics},
  url       = {https://aclanthology.org/2025.acl-long.1021/},
  doi       = {10.18653/v1/2025.acl-long.1021},
  pages     = {20959--20998},
  isbn      = {979-8-89176-251-0}
}

@inproceedings{yoon-etal-2024-evaluating,
  title     = {Evaluating Large Language Models as Generative User Simulators for Conversational Recommendation},
  author    = {Yoon, Se-eun and He, Zhankui and Echterhoff, Jessica and McAuley, Julian},
  editor    = {Duh, Kevin and Gomez, Helena and Bethard, Steven},
  booktitle = {Proceedings of the 2024 Conference of the North American Chapter of the Association for Computational Linguistics: Human Language Technologies (Volume 1: Long Papers)},
  month     = jun,
  year      = {2024},
  address   = {Mexico City, Mexico},
  publisher = {Association for Computational Linguistics},
  url       = {https://aclanthology.org/2024.naacl-long.83/},
  doi       = {10.18653/v1/2024.naacl-long.83},
  pages     = {1490--1504}
}

@inproceedings{zelikman-etal-2023-generating,
  title     = {Generating and Evaluating Tests for K-12 Students with Language Model Simulations: A Case Study on Sentence Reading Efficiency},
  author    = {Zelikman, Eric and Ma, Wanjing and Tran, Jasmine and Yang, Diyi and Yeatman, Jason and Haber, Nick},
  editor    = {Bouamor, Houda and Pino, Juan and Bali, Kalika},
  booktitle = {Proceedings of the 2023 Conference on Empirical Methods in Natural Language Processing},
  month     = dec,
  year      = {2023},
  address   = {Singapore},
  publisher = {Association for Computational Linguistics},
  url       = {https://aclanthology.org/2023.emnlp-main.135/},
  doi       = {10.18653/v1/2023.emnlp-main.135},
  pages     = {2190--2205}
}

\clearpage
\appendix
\section{Reading Comprehension Dataset}
\label{app:dataset}

This section provides additional details regarding the digital platform used for data collection, the distribution of student responses, and our dataset filtering procedures.

\begin{figure}[!th]
  \centering
  \includegraphics[width=\linewidth]{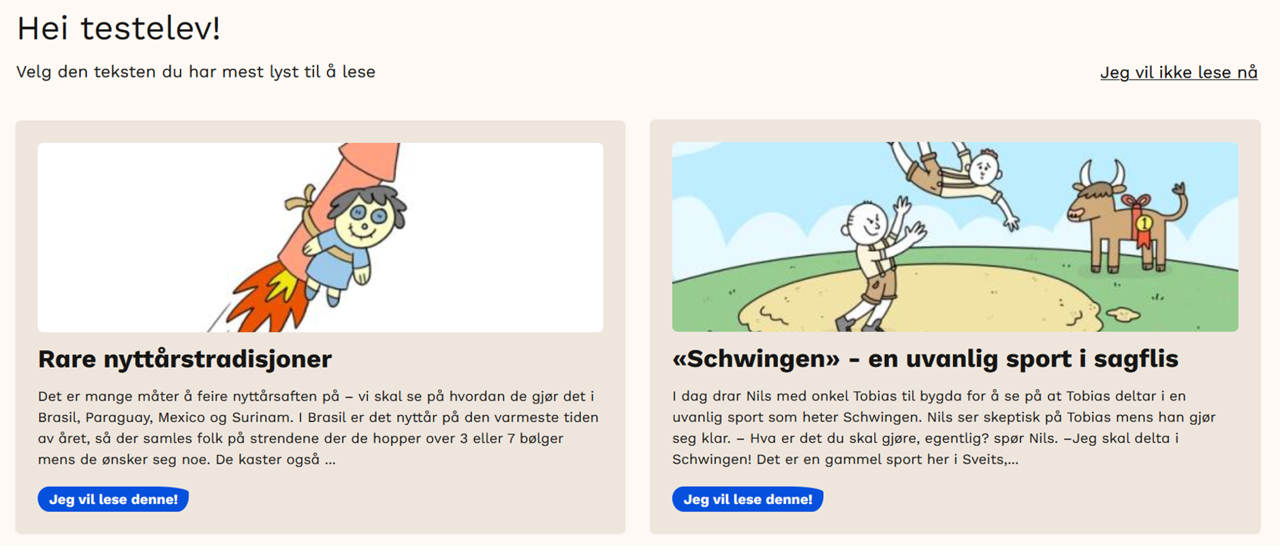}\\[1em]
  \includegraphics[width=\linewidth]{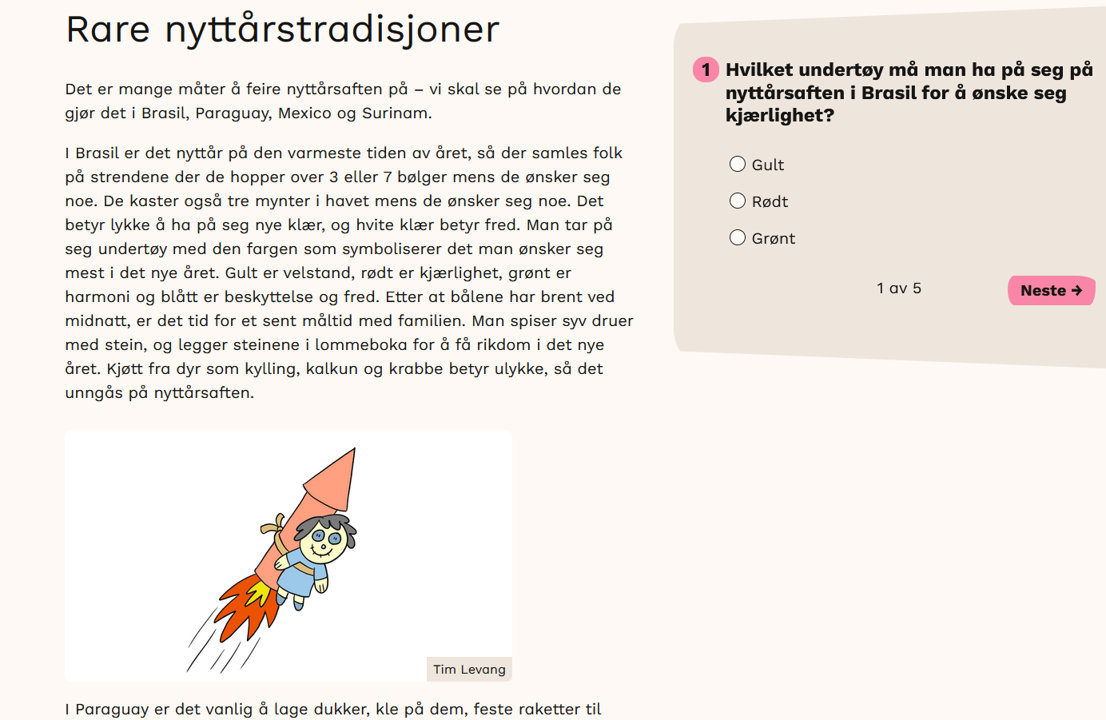}
  \caption{User interface of the reading comprehension platform. Top: The text selection screen where students choose between two randomly paired texts (English translation: (Left) ``Unusual New Year's Traditions,'' (Right) ``Schwingen - an Unusual Sport on Sawdust.'' Texts: Kirsti Thisland and Michelle N. Maurer; Illustrations: Tim Levang). Bottom: The assessment interface where students answer questions based on the text (English translation: ``What underwear should you wear in Brazil on New Year's Eve to wish for love?'', options: ``Yellow,'' ``Red,'' ``Green.'' Text: Kirsti Thisland; Illustration: Tim Levang).}
  \label{fig:ui_screenshots}
\end{figure}

\paragraph{User Interface and Text Selection}
The empirical data was collected via a digital educational platform designed for Norwegian 4th--6th-grade students. As illustrated in Figure~\ref{fig:ui_screenshots}, the workflow begins with a text selection screen (Top). Students are presented with a choice between two randomly paired texts to encourage engagement. After making their selection and reading the material, students proceed to an assessment interface where they answer a series of comprehension questions (Bottom).

\paragraph{Data Filtering and Modality Constraints}
Because our user simulators are powered by text-based Large Language Models, it is critical that the comprehension questions do not rely on unprovided visual context. We manually verified all texts and questions to ensure they are fully answerable relying exclusively on the provided reading material and any associated image alt text. During this review process, we identified two texts that strictly required visual inspection of accompanying images to successfully answer the questions. To maintain a fair evaluation environment for the text-based LLMs, these two texts, along with their associated questions and student responses, were entirely removed from the final dataset.

\begin{figure}[!thbp]
  \centering
  \vspace{-0.5\baselineskip}
  \includegraphics[width=0.9\linewidth]{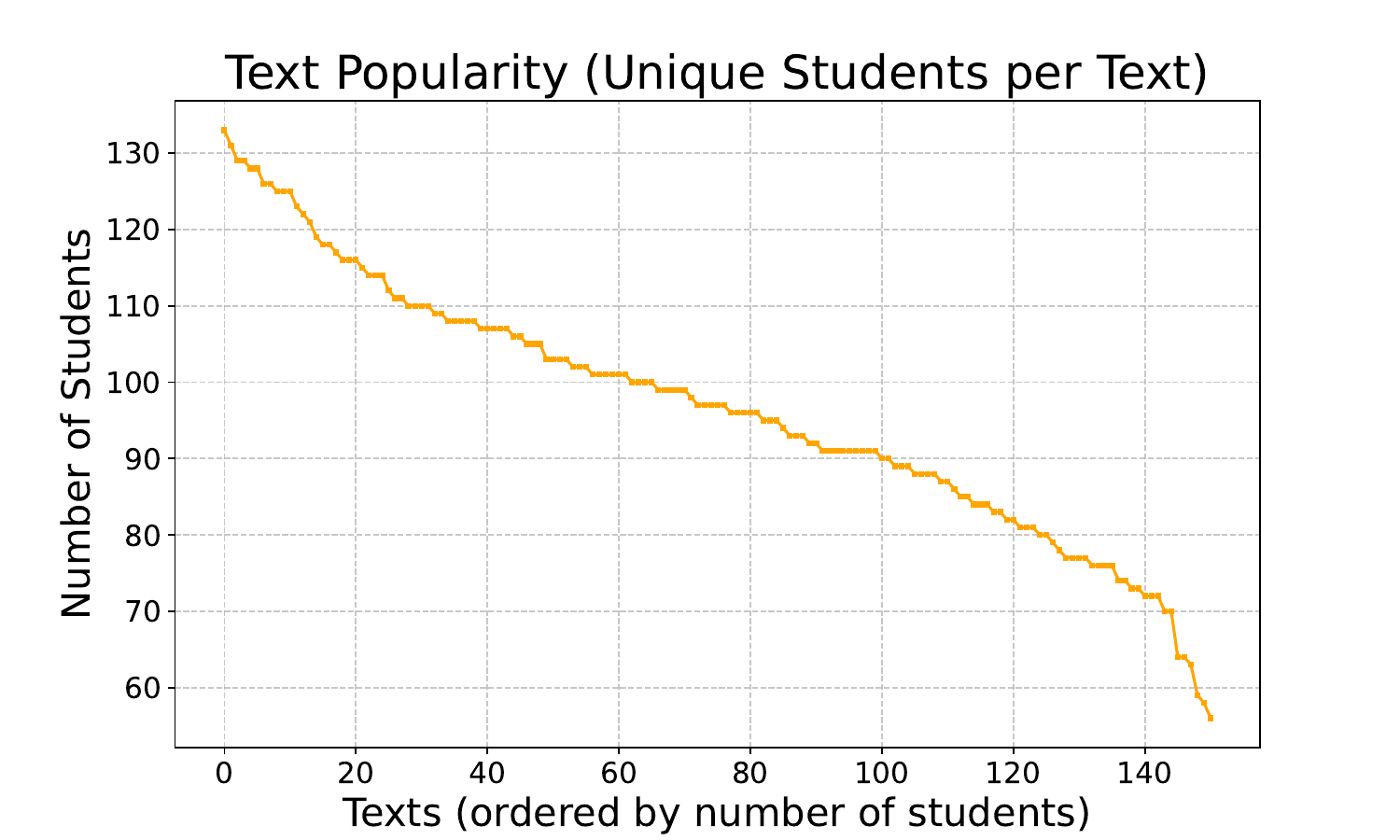}
  \caption{Text popularity distribution, showing the number of students who selected and completed the assessment for each text, ordered from most to least popular.}
  \label{fig:text_popularity}
\end{figure}

\paragraph{Text Popularity Distribution}
Because the platform allows students to choose between two text options, the distribution of responses is not uniform across the dataset. Figure~\ref{fig:text_popularity} illustrates the popularity of the texts, measured by the number of unique students who selected them. The distribution shows a relatively smooth, steady decline across the dataset; the most frequently chosen texts accumulated over 140 student responses, while the least selected texts received approx. 55 responses.

\section{Question Categorization Methodology}
\label{app:question_categories}

To classify the 750 questions into the three cognitive categories (Locate, Interpret and Connect, and Reflect and Evaluate), we employed a semi-automatic pipeline utilizing an ensemble of three distinct LLMs: Gemini-3.5-Flash-preview, GPT-5.4, and Llama-3.3-70B Instruct.

Each model was provided with a zero-shot prompt outlining the category definitions, disambiguation tips, and output formatting instructions (see Figure~\ref{fig:prompt_categorization}). We adopted a strict consensus-based approach for the initial labeling. If all three models predicted the same category, the label was provisionally accepted. In cases where there was any disagreement among the models' predictions—which occurred for 83 out of the 750 questions—the item was flagged for manual review. All 83 flagged questions were manually checked and resolved by an expert specialized in the development of reading comprehension questions.

To validate the accuracy of the automated consensus, the expert also randomly sampled and reviewed an additional 18 questions where all three LLMs had achieved perfect agreement. In all 18 validation cases, the expert's manual assessment matched the LLM consensus, confirming the reliability of the ensemble approach.

\begin{figure}[!htbp]
  \centering
  \fbox{
    \begin{minipage}{0.95\linewidth}
      \scriptsize
      \ttfamily
      You are an expert educational assessor and psychometrician specializing
      in reading comprehension. Your task is to analyze a reading
      comprehension text and a specific question, and classify the question
      based on the cognitive process it targets.\\[1em]
      Here are the three categories and their definitions:\\[1em]
      1. **Locate**: Focuses on finding and retrieving information. This
      involves reading comprehension processes such as finding information
      that is explicitly stated in one or several places in a text, locating
      competing information, and finding information that is implicitly
      expressed in a localized part of the text.\\
      2. **Interpret and Connect**: Focuses on synthesizing and concluding.
      This involves reading comprehension processes such as drawing
      conclusions based on information from several different places within
      a text, or across multiple texts.\\
      3. **Reflect and Evaluate**: Focuses on critical thinking and holistic
      understanding. This involves reading comprehension processes such as
      engaging independently and critically with a text, understanding the
      text's overall meaning/main idea, and commenting on or justifying
      one's own viewpoints, analyses, and/or evaluations regarding the
      text's content, form, or author's intent.\\[1em]
      \#\#\# Tips for Disambiguation:\\
      * If the student just needs to "put their finger on it" (even if they
      have to sift through competing facts or read between the lines of a
      specific sentence), it is **Locate**.\\
      * If the student must stitch together "Fact A" from paragraph 1 and
      "Fact B" from paragraph 4 to figure out "Conclusion C", it is
      **Interpret and Connect**.\\
      * If the student is asked to identify the overarching theme, judge the
      reliability of the author, apply the text to the real world, or give
      an opinion supported by the text, it is **Reflect and Evaluate**.\\[1em]
      \#\#\# Language Note:\\
      The reading text and question are in Norwegian. Please analyze the
      text and question in their original Norwegian context to capture the
      correct nuances, but write your analysis and output entirely in
      English.\\[1em]
      \#\#\# Input:\\
      **Text:**\\
      \textcolor{blue}{\{text\_content\}}\\[1em]
      **Question:**\\
      \textcolor{blue}{\{question\}}\\[1em]
      \#\#\# Output Format:\\
      Please provide your response in the following format:\\
      **Step 1: Cognitive Task Analysis:** Briefly explain exactly what the
      reader must do to arrive at the answer. (e.g., "The reader must look
      at paragraph 2 and paragraph 5 and synthesize the two events to
      understand why the character is sad.")\\
      **Step 2: Category Match:** Explain which category best aligns with
      the task analysis.\\
      **Step 3: Final Classification:** Provide ONLY the name of the exact
      category (Locate, Interpret and Connect, or Reflect and Evaluate).
    \end{minipage}
  }
  \caption{Zero-shot prompt used for automatic question categorization.}
  \label{fig:prompt_categorization}
\end{figure}

\begin{figure}[!thbp]
  \centering
  \includegraphics[width=0.9\linewidth]{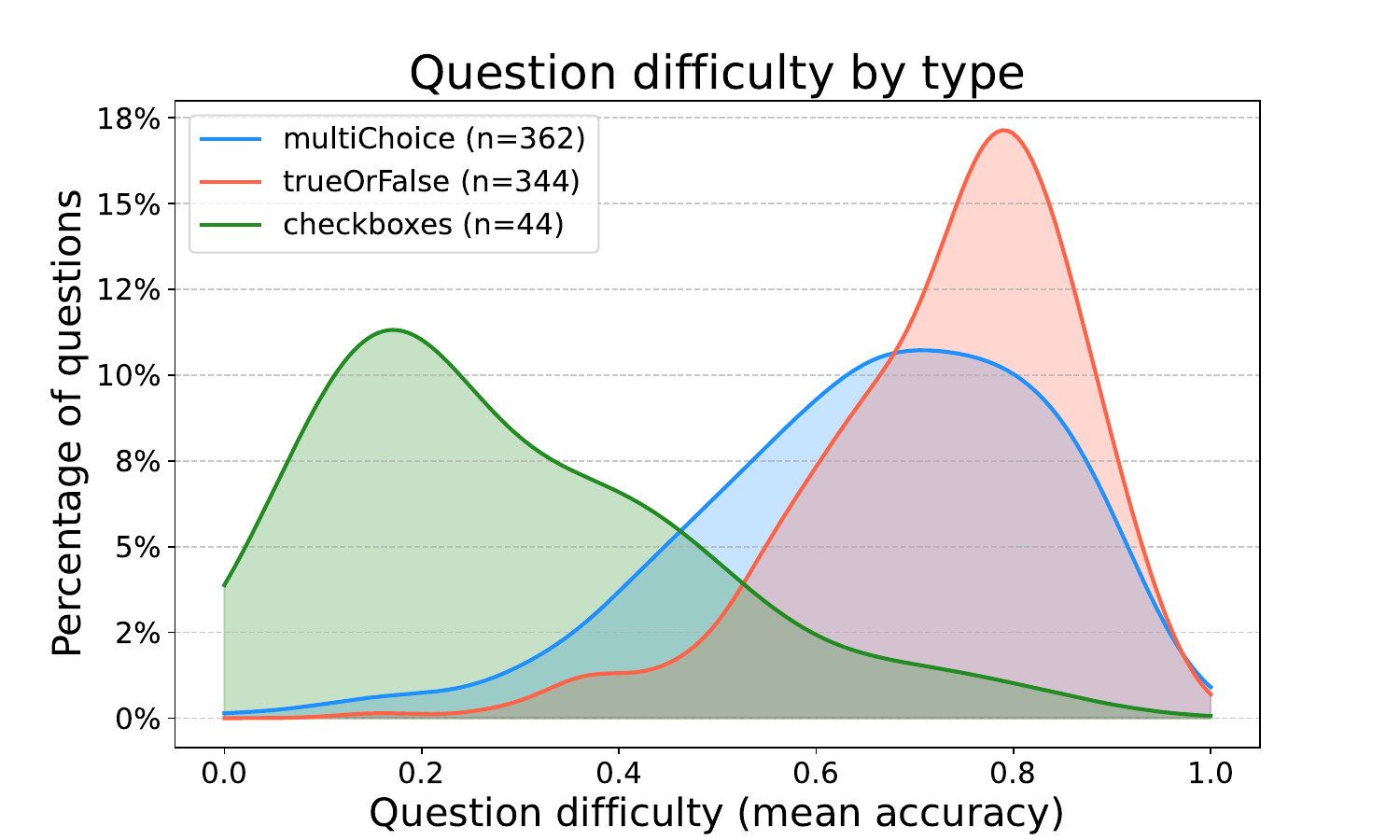}\\[0.5em]
  \includegraphics[width=0.9\linewidth]{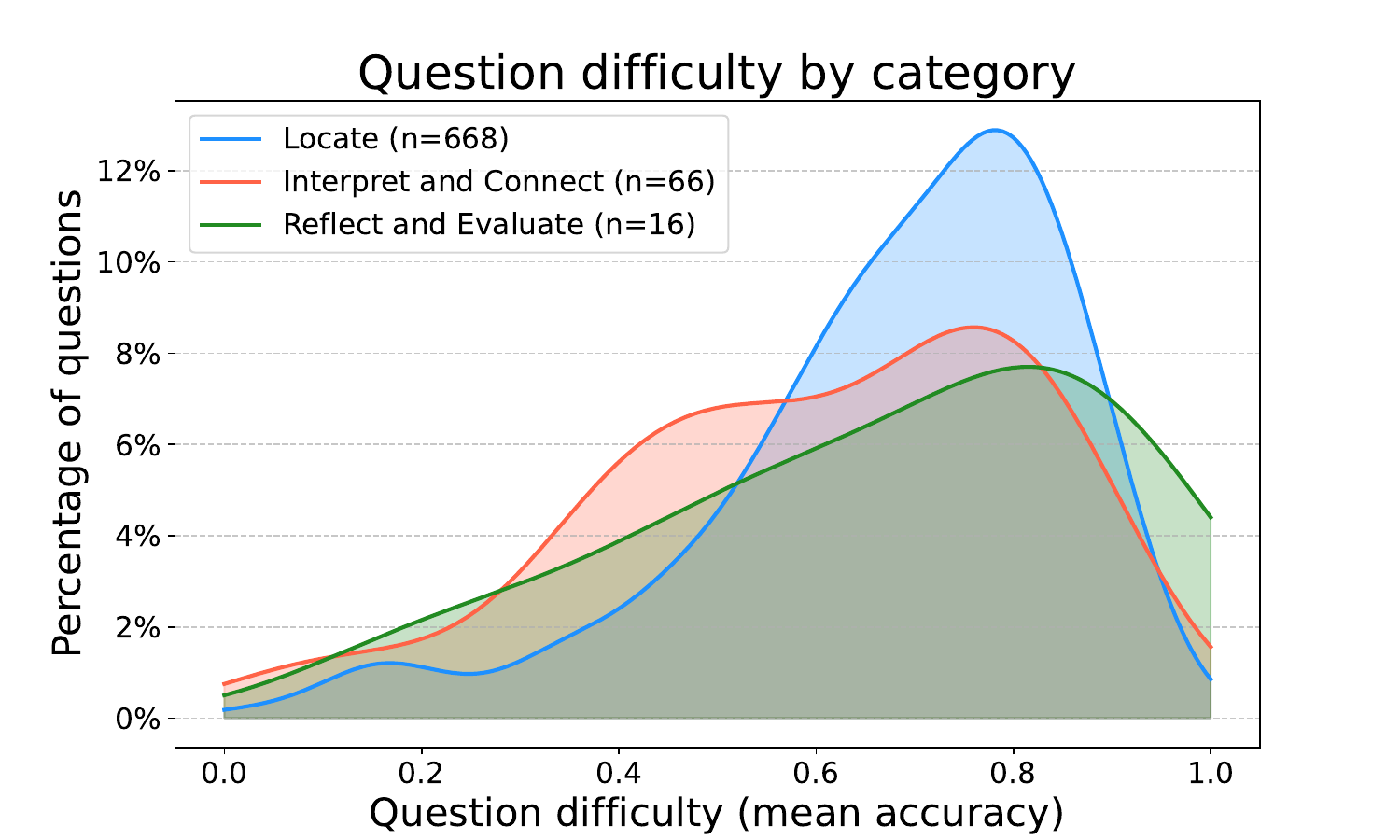}  \caption{Empirical question difficulty (measured as mean student accuracy). The top plot breaks down difficulty by structural question type ($n=362$ multiChoice, $n=344$ trueOrFalse, $n=44$ checkboxes). The bottom plot breaks down difficulty by pedagogical category ($n=668$ Locate, $n=66$ Interpret \& Connect, $n=16$ Reflect \& Evaluate).}
  \label{fig:difficulty_distributions}
\end{figure}

\subsection{Empirical Question Difficulty Distributions}
\label{app:question_difficulty}

To better understand the composition and challenge level of our evaluation dataset, we analyze the empirical difficulty of the 750 questions. We formally define the difficulty of a given question $q$ as its mean accuracy across the human population. Specifically, this is calculated as the average of the binary student responses $(s, q) \in \{0, 1\}$ over all students $s$ who answered that specific question.

Figure~\ref{fig:difficulty_distributions} presents empirical question difficulty across the dataset. The left plot visualizes the distribution split by structural question type. Multi-select checkbox questions are demonstrably more difficult for the student population (peaking near 0.2 mean accuracy) compared to single-select multiple-choice and true-or-false formats, which peak between 0.7 and 0.8. The right plot visualizes the distribution split by pedagogical category. While ``Locate'' questions tend to have high success rates, questions requiring higher-order cognitive skills, such as ``Interpret \& Connect'' and ``Reflect \& Evaluate,'' exhibit wider variance and generally lower mean accuracies.

\section{Administrative and Ethical Details}
\label{app:ethics}

As noted in Section~\ref{sec:expsetup}, this study utilizes data from the Norwegian primary education system. To recruit participants, the research team directly contacted several schools and published a call for participation on social media sites followed by Norwegian primary school teachers. This initially resulted in 151 classes from 4th to 6th grade (ages 8--11) registering for the reading assessment. Note that some attrition occurred prior to data collection, primarily due to student illnesses and busy school schedules.

Given the strictly anonymous nature of the secondary data analysis and the privacy-preserving default of the data collection (where no personally identifiable information was recorded and individual identifiers are not linkable to real-world identities), formal evaluation by a national agency (SIKT, the Data Protection Services of the Norwegian Agency for Shared Services in Education and Research) for research purposes was not required.

Regarding the data collection session structure, teachers who registered their class organized the assessment locally based on a detailed letter of instruction. Students were asked to sit for a total of 45 minutes. This included an initial briefing by their teacher, which was estimated to take approximately 5 minutes. The students then completed as many reading comprehension texts and question sets as they could in the remaining time, resulting in an overall average of around 7 texts completed per student.

\section{Question-Specific Prompt Formatting}
\label{app:question_prompts}

As described in Section~\ref{sec:baseline}, the baseline persona simulator utilizes a primary prompt template where the \texttt{\{question\_text\}} placeholder is dynamically populated based on the format of the comprehension question. To ensure automated, standardized extraction of the simulated student's answers without relying on brittle regex parsing, we append strict JSON formatting instructions to each question type.

The exact prompt elements injected for True/False, Multiple-Choice, and Checkbox questions are detailed in Figures~\ref{fig:prompt_tf}, \ref{fig:prompt_mc}, and \ref{fig:prompt_cb}, respectively.

\begin{figure}[!htbp]
  \centering
  \fbox{
    \begin{minipage}{0.95\linewidth}
      \scriptsize
      \ttfamily
      This is a true or false question. Decide whether the
      following statement is true or false based on the
      text.\\[1em]
      Statement: \textcolor{blue}{\{statement\}}\\[1em]
      Respond with ONLY a JSON object in the following
      format, with no other text:
      \{"answer": true\}
      or
      \{"answer": false\}
    \end{minipage}
  }
  \caption{Prompt element injected into \texttt{\{question\_text\}} for True/False questions.}
  \label{fig:prompt_tf}
\end{figure}

\begin{figure}[!htbp]
  \centering
  \fbox{
    \begin{minipage}{0.95\linewidth}
      \scriptsize
      \ttfamily
      This is a multiple choice question. Choose exactly ONE
      answer from the options below.\\[1em]
      Options:
      \textcolor{blue}{\{options\_text\}}\\[1em]
      Respond with ONLY a JSON object in the following
      format, with no other text:
      \{"answer": "<letter>"\}
      where <letter> is the letter (A, B, C, ...) of your
      chosen option.
    \end{minipage}
  }
  \caption{Prompt element injected into \texttt{\{question\_text\}} for Multiple-Choice questions.}
  \label{fig:prompt_mc}
\end{figure}

\begin{figure}[!htbp]
  \centering
  \fbox{
    \begin{minipage}{0.95\linewidth}
      \scriptsize
      \ttfamily
      This is a checkbox question. For each option below,
      decide whether it is correct or not.\\[1em]
      Options:
      \textcolor{blue}{\{options\_text\}}\\[1em]
      Respond with ONLY a JSON object in the following
      format, with no other text:
      \{"answer": [true, false, ...]\}\\
      The list must contain one true/false value for each
      option, in the same order as listed above.
    \end{minipage}
  }
  \caption{Prompt element injected into \texttt{\{question\_text\}} for Checkbox (multi-select) questions.}
  \label{fig:prompt_cb}
\end{figure}

Finally, Figure~\ref{fig:baseline_prompt_no} shows the fully Norwegian (Bokm{\aa}l) variant of the baseline persona prompt, used for the instruction-language robustness check reported with the pilot baselines (Table~\ref{tab:pilot_full}).

\begin{figure}[!htbp]
  \centering
  \fbox{
    \begin{minipage}{0.95\linewidth}
      \scriptsize
      \ttfamily
      Du simulerer en elev på 5. trinn i Norge som tar en leseforståelsesprøve. Les teksten nedenfor og svar på spørsmålet slik en typisk 10--11 år gammel elev ville gjort. Eleven svarer ikke alltid riktig.\\[1em]
      <TEXT>
      \textcolor{blue}{\{text\_content\}}
      </TEXT>\\[1em]
      <QUESTION>
      \textcolor{blue}{\{question\_text\}}
      </QUESTION>
    \end{minipage}
  }
  \caption{Fully Norwegian (Bokm{\aa}l) variant of the baseline persona prompt (cf.\ the English template in Figure~\ref{fig:baseline_prompt}), used for the Norwegian-prompt robustness check to test whether the simulation gap is an instruction-language artifact rather than a genuine cognitive limitation. Per-question formatting mirrors the English templates above (Figures~\ref{fig:prompt_tf}--\ref{fig:prompt_cb}), translated to Bokm{\aa}l.}
  \label{fig:baseline_prompt_no}
\end{figure}

\section{Formal Definitions of Evaluation Metrics}
\label{app:metrics}

This section provides the formal mathematical definitions for the evaluation metrics introduced in Section~\ref{sec:exp_setup:metrics}. Throughout these definitions, let $r_{s,q} \in \{0,1\}$ represent the discrete, binary response of a subject $s$ (either a real student or a simulated agent) on a specific question $q$.

\subsection{Student-Centric Metrics}

These metrics assess how well the simulated agents approximate the overall performance, variance, and calibration of the aggregate human student population.

\paragraph{Absolute Performance Gap}
To evaluate the macro-level accuracy of the simulators, we compute the overall mean success rate of a given population. Let $R$ denote the total set of valid response tuples $(s,q)$ for a population. The mean population accuracy $\mu$ is defined as:
$$\mu = \frac{1}{|R|} \sum_{(s,q) \in R} r_{s,q}$$
The Absolute Performance Gap is simply the absolute difference between the human and simulated population means: $|\mu_{Human} - \mu_{Sim}|$.

\paragraph{Distributional Alignment (JSD)}
To ensure the simulators capture the natural variance of the student population rather than collapsing to a deterministic mean, we measure the distance between their score distributions. Let a subject's overall score be the average of their responses: $c_s = \frac{1}{|Q_s|} \sum_{q \in Q_s} r_{s,q}$, where $Q_s$ is the set of questions answered by subject $s$. We estimate the discrete probability distributions of these scores for the human population ($P$) and the simulated population ($Q$). The alignment is calculated using the symmetric Jensen-Shannon Divergence (JSD).

\paragraph{Expected Calibration Error (ECE)}
While the absolute performance gap provides a global evaluation, it can mask non-uniform miscalibration---for instance, if a simulator systematically underestimates performance on challenging questions while overestimating it on moderately difficult ones. To quantify this, we adapt the Expected Calibration Error (ECE) metric to measure the alignment between simulated and human success rates across varying levels of question difficulty~\citep{seshadri-etal-2026-lost}.

We partition the questions into $M=6$ discrete difficulty bins based on empirical success rates. Let $S_i^{(Human)}$ and $S_i^{(Sim)}$ denote the average success rates of the human and simulated populations for questions within bin $i$, respectively. Let $w_i$ denote the proportion of total questions that fall into bin $i$. The calibration error is then computed as the weighted average absolute deviation across all difficulty bins:
$$ECE = \sum_{i=1}^{M} w_i |S_i^{(Human)} - S_i^{(Sim)}|$$
A perfectly calibrated simulator will yield an ECE of $0$, indicating that its predicted performance identically matches human outcomes across every difficulty stratum~\citep{seshadri-etal-2026-lost}.

\subsection{Item-Centric Metrics}

These metrics isolate performance on individual questions to evaluate whether simulators struggle with the exact same linguistic and cognitive challenges as developing readers.

\paragraph{Item-Level Difficulty Correlation}
We formally define the empirical difficulty of a specific question $q$, denoted as $d_q$, as its mean success rate across all subjects who answered it:
$$d_q = \frac{1}{|S_q|} \sum_{s \in S_q} r_{s,q}$$
where $S_q$ is the set of subjects exposed to question $q$. Let $D^{(Human)}$ and $D^{(Sim)}$ be the corresponding vectors of question difficulties for the human and simulated populations. To evaluate item-level alignment, we compute both the Pearson correlation coefficient ($\rho$) to measure the linear relationship and Spearman's rank correlation coefficient ($r_s$) to measure the monotonic rank-order relationship between $D^{(Human)}$ and $D^{(Sim)}$.

\section{Model Providers and API Details}
\label{app:model_details}

All inference was routed through the OpenRouter API. Table~\ref{tab:model_details} lists the exact providers, human-readable model names referenced throughout this paper, and the corresponding API endpoint snapshots used in the experiments.

\begin{table*}[ht]
  \caption{Detailed mapping of the LLMs evaluated in our study, including the specific model snapshot tags queried via the OpenRouter API.}
  \label{tab:model_details}
  \centering
  \small
  \renewcommand{\arraystretch}{1.2}
  \begin{tabular}{lll}
    \toprule
    \textbf{Name}          & \textbf{Provider} & \textbf{OpenRouter API Endpoint / Snapshot} \\
    \midrule
    Llama-3.3-70B-Instruct & Meta              & \texttt{meta-llama/llama-3.3-70b-instruct}  \\
    Mixtral-8x22B-Instruct & Mistral AI        & \texttt{mistralai/mixtral-8x22b-instruct}   \\
    GPT-4o-mini            & OpenAI            & \texttt{openai/gpt-4o-mini}                 \\
    GPT-5.4                & OpenAI            & \texttt{openai/gpt-5.4}                     \\
    Gemini-3.5-Flash-Lite  & Google            & \texttt{google/gemini-3.5-flash-lite}       \\
    Gemini-3.7-Flash       & Google            & \texttt{google/gemini-3.7-flash}            \\
    \bottomrule
  \end{tabular}
\end{table*}

\section{Full Pilot Results with Run Variance}
\label{app:pilot_full}

Table~\ref{tab:pilot_full} reports the complete metric set for the pilot baselines (Section~\ref{sec:pilot_results}) as mean $\pm$ standard deviation over the $n=3$ runs; the main pilot table (Table~\ref{tab:pilot_results}) reports Mean Accuracy and JSD only for readability. Run-to-run variance is small across all backbones, confirming that the superhuman bias is a stable property of persona prompting rather than a sampling artifact.

\begin{table*}[ht]
  \caption{Full pilot results (baseline persona-based prompting), reported as mean $\pm$ standard deviation over $n=3$ runs. Real students (Held-out) is a single population, so no run variance is reported. The two \emph{Norwegian} rows apply the fully Norwegian (Bokm{\aa}l) prompt variant (Figure~\ref{fig:baseline_prompt_no}) to the two selected backbones.}
  \label{tab:pilot_full}
  \centering
  \scriptsize
  \begin{tabular}{l r@{\,}l r@{\,}l r@{\,}l r@{\,}l r@{\,}l r@{\,}l}
    \toprule
                                       & \multicolumn{8}{c}{\textbf{Student-Centric}} & \multicolumn{4}{c}{\textbf{Item-Centric}}                                                                                                                                                                                                                                                                                                           \\
    \cmidrule(lr){2-9} \cmidrule(lr){10-13}
    \textbf{Population}                & \multicolumn{2}{c}{\textbf{Mean Acc.}}       & \multicolumn{2}{c}{\textbf{Abs. Gap} ($\downarrow$)} & \multicolumn{2}{c}{\textbf{JSD} ($\downarrow$)} & \multicolumn{2}{c}{\textbf{ECE} ($\downarrow$)} & \multicolumn{2}{c}{\textbf{Pearson $\rho$} ($\uparrow$)} & \multicolumn{2}{c}{\textbf{Spearman $r_s$} ($\uparrow$)}                                                                      \\
    \midrule
    Real students (Held-out)           & 0.661                                        &                                                      & 0.026                                           &                                                 & 0.006                                                    &                                                          & 0.027 &              & 0.895 &              & 0.845 &              \\
    \midrule
    Random Guessing                    & 0.378                                        & \stdev{.001}                                         & 0.309                                           & \stdev{.001}                                    & 0.518                                                    & \stdev{.004}                                             & 0.306 & \stdev{.001} & 0.409 & \stdev{.014} & 0.319 & \stdev{.008} \\
    \hdashline
    Llama-3.3-70B-Instruct             & 0.966                                        & \stdev{.001}                                         & 0.279                                           & \stdev{.001}                                    & 0.679                                                    & \stdev{.011}                                             & 0.285 & \stdev{.001} & 0.241 & \stdev{.003} & 0.179 & \stdev{.045} \\
    Mixtral-8x22B-Instruct             & 0.956                                        & \stdev{.000}                                         & 0.269                                           & \stdev{.000}                                    & 0.634                                                    & \stdev{.003}                                             & 0.276 & \stdev{.000} & 0.326 & \stdev{.002} & 0.289 & \stdev{.020} \\
    GPT-4o-mini                        & 0.958                                        & \stdev{.001}                                         & 0.271                                           & \stdev{.001}                                    & 0.646                                                    & \stdev{.002}                                             & 0.278 & \stdev{.000} & 0.315 & \stdev{.002} & 0.240 & \stdev{.008} \\
    GPT-5.4                            & 0.974                                        & \stdev{.000}                                         & 0.287                                           & \stdev{.000}                                    & 0.732                                                    & \stdev{.001}                                             & 0.293 & \stdev{.000} & 0.255 & \stdev{.001} & 0.172 & \stdev{.002} \\
    Gemini-3.5-Flash-Lite              & 0.972                                        & \stdev{.000}                                         & 0.285                                           & \stdev{.000}                                    & 0.713                                                    & \stdev{.004}                                             & 0.291 & \stdev{.000} & 0.290 & \stdev{.001} & 0.220 & \stdev{.008} \\
    Gemini-3.7-Flash                   & 0.985                                        & \stdev{.000}                                         & 0.298                                           & \stdev{.000}                                    & 0.773                                                    & \stdev{.001}                                             & 0.305 & \stdev{.000} & 0.246 & \stdev{.001} & 0.154 & \stdev{.006} \\
    \hdashline
    Llama-3.3-70B-Instruct (Norwegian) & 0.963                                        & \stdev{.001}                                         & 0.276                                           & \stdev{.001}                                    & 0.666                                                    & \stdev{.002}                                             & 0.283 & \stdev{.001} & 0.277 & \stdev{.010} & 0.196 & \stdev{.014} \\
    Gemini-3.5-Flash-Lite (Norwegian)  & 0.969                                        & \stdev{.000}                                         & 0.282                                           & \stdev{.000}                                    & 0.696                                                    & \stdev{.007}                                             & 0.288 & \stdev{.000} & 0.335 & \stdev{.006} & 0.269 & \stdev{.007} \\
    \bottomrule
  \end{tabular}
\end{table*}

\section{Diagnostic Baselines and Ablation}
\label{app:diagnostics}

To understand \emph{why} CBUS improves fidelity---and to rule out simpler explanations---we evaluate three diagnostic baselines on two backbones (Llama-3.3-70B and Gemini-3.5-Flash-Lite), all under the same protocol as our main experiments (Table~\ref{tab:diagnostics}). The first two test whether merely matching the human mean accuracy is sufficient; the third ablates the salience-based selection inside CBUS-SPR. A complementary check on instruction language (fully Norwegian prompts) is reported with the pilot baselines in Table~\ref{tab:pilot_full}.

\paragraph{Calibrated Low-persona}
The most direct way to lower a simulator's accuracy is to ask for it. If explicitly instructing the model to behave like a weak student reproduced empirical behavior, no architectural mechanism would be required.\\
\emph{Implementation.} This is the standard persona-prompted baseline with an accuracy-calibrated prompt: the model is told it is a ``below-average'' reader who ``answers only about two out of every three questions correctly (roughly 65--70\% accuracy)'' and should make ``realistic mistakes on the harder questions.'' No other part of the method changes.

\paragraph{Answer-noising}
This baseline forces the mean accuracy to match the human population \emph{exactly}, testing whether hitting the right mean also reproduces the score distribution and the item-difficulty structure. \\
\emph{Implementation.} It is a post-hoc transformation of the persona-baseline outputs and makes no additional LLM calls. Given a baseline run with mean accuracy $a_0$ and a target $a^\star$ (the real-student mean, $0.687$), each currently-correct response is independently flipped to incorrect with probability $(a_0 - a^\star)/a_0$, so the expected accuracy after noising equals $a^\star$. Flips are type-appropriate (negating a true/false answer, selecting a wrong option, or breaking a checkbox match) and seeded per run for reproducibility. Crucially, because flips are applied uniformly at random---independent of which questions are genuinely difficult---this baseline carries no item-difficulty signal by construction.

\paragraph{Random Proposition Dropout}
CBUS-SPR could help for two distinct reasons: the \emph{capacity bound} (only $C$ propositions are retained) or the \emph{salience-based selection} (the model keeps the $C$ it judges most important). This ablation separates the two by holding the bound fixed while removing the selection.\\
\emph{Implementation.} It is identical to CBUS-SPR except in the Stage-1 encoding step: rather than extracting exactly $C$ propositions, the model extracts a larger salient pool of $3 \times C=12$ propositions, from which $C$ are then retained \emph{uniformly at random} instead of most-salient-first. Stage 2 (answering from only the retained propositions) is unchanged, so the amount remembered is identical to CBUS-SPR; only \emph{which} propositions survive differs. Note that the discarded propositions are themselves drawn from a salient pool, so this ablates fine-grained ranking \emph{within} salient content rather than salient-versus-arbitrary selection.

\begin{table*}[t]
  \caption{Diagnostic baselines and ablation on two backbones, under independent per-student sampling, reported as mean $\pm$ standard deviation over $n=3$ runs.}
  \label{tab:diagnostics}
  \centering
  \scriptsize
  \begin{tabular}{ll r@{\,}l r@{\,}l r@{\,}l r@{\,}l r@{\,}l}
    \toprule
                                                 &                        & \multicolumn{6}{c}{\textbf{Student-Centric}}         & \multicolumn{4}{c}{\textbf{Item-Centric}}                                                                                                                                                                                                                                            \\
    \cmidrule(lr){3-8} \cmidrule(lr){9-12}
    \textbf{Backbone}                            & \textbf{Variant}       & \multicolumn{2}{c}{\textbf{Abs. Gap} ($\downarrow$)} & \multicolumn{2}{c}{\textbf{JSD} ($\downarrow$)} & \multicolumn{2}{c}{\textbf{ECE} ($\downarrow$)} & \multicolumn{2}{c}{\textbf{Pearson $\rho$} ($\uparrow$)} & \multicolumn{2}{c}{\textbf{Spearman $r_s$} ($\uparrow$)}                                                              \\
    \midrule
    \multicolumn{2}{l}{Real students (Held-out)} & 0.026                  &                                                      & 0.006                                           &                                                 & 0.027                                                    &                                                          & 0.895        &       & 0.845        &                      \\
    \midrule
    Llama-3.3-70b-Instruct                       & baseline               & 0.279                                                & \stdev{.001}                                    & 0.679                                           & \stdev{.011}                                             & 0.285                                                    & \stdev{.001} & 0.241 & \stdev{.003} & 0.179 & \stdev{.045} \\
                                                 & Calibrated low-persona & 0.277                                                & \stdev{.001}                                    & 0.661                                           & \stdev{.003}                                             & 0.284                                                    & \stdev{.001} & 0.220 & \stdev{.004} & 0.210 & \stdev{.009} \\
                                                 & Answer-noising         & 0.001                                                & \stdev{.001}                                    & 0.102                                           & \stdev{.005}                                             & 0.130                                                    & \stdev{.003} & 0.216 & \stdev{.026} & 0.065 & \stdev{.040} \\
                                                 & Random dropout         & 0.004                                                & \stdev{.001}                                    & 0.084                                           & \stdev{.007}                                             & 0.092                                                    & \stdev{.002} & 0.304 & \stdev{.001} & 0.266 & \stdev{.002} \\
                                                 & CBUS-SPR               & 0.037                                                & \stdev{.002}                                    & 0.093                                           & \stdev{.007}                                             & 0.069                                                    & \stdev{.002} & 0.301 & \stdev{.007} & 0.297 & \stdev{.004} \\
                                                 & CBUS-TS                & 0.238                                                & \stdev{.001}                                    & 0.475                                           & \stdev{.007}                                             & 0.244                                                    & \stdev{.001} & 0.455 & \stdev{.003} & 0.304 & \stdev{.016} \\
    \midrule
    Gemini-3.5-Flash-Lite                        & baseline               & 0.285                                                & \stdev{.000}                                    & 0.713                                           & \stdev{.004}                                             & 0.291                                                    & \stdev{.000} & 0.290 & \stdev{.001} & 0.220 & \stdev{.008} \\
                                                 & Calibrated low-persona & 0.073                                                & \stdev{.001}                                    & 0.106                                           & \stdev{.001}                                             & 0.082                                                    & \stdev{.000} & 0.340 & \stdev{.002} & 0.320 & \stdev{.000} \\
                                                 & Answer-noising         & 0.001                                                & \stdev{.001}                                    & 0.093                                           & \stdev{.013}                                             & 0.130                                                    & \stdev{.002} & 0.266 & \stdev{.015} & 0.090 & \stdev{.032} \\
                                                 & Random dropout         & 0.030                                                & \stdev{.001}                                    & 0.100                                           & \stdev{.001}                                             & 0.096                                                    & \stdev{.002} & 0.316 & \stdev{.004} & 0.260 & \stdev{.006} \\
                                                 & CBUS-SPR               & 0.031                                                & \stdev{.002}                                    & 0.086                                           & \stdev{.004}                                             & 0.074                                                    & \stdev{.001} & 0.304 & \stdev{.001} & 0.295 & \stdev{.007} \\
                                                 & CBUS-TS                & 0.238                                                & \stdev{.001}                                    & 0.470                                           & \stdev{.007}                                             & 0.243                                                    & \stdev{.001} & 0.454 & \stdev{.002} & 0.298 & \stdev{.009} \\
    \bottomrule
  \end{tabular}
\end{table*}

\paragraph{Results} The results (Table~\ref{tab:diagnostics}) support two conclusions. First, matching---or merely attempting to match---the human accuracy is not sufficient for fidelity. \emph{Answer-noising} forces the mean to match exactly (Abs.\ Gap of $0.001$ on both backbones), yet its item-difficulty correlation nearly vanishes (Spearman $r_s$ of $0.065$ and $0.090$ for Llama-3.3-70B and Gemini-3.5-Flash-Lite), because uniform random flipping carries no signal about which questions are hard. The \emph{Calibrated low-persona} baseline instead prompts the model to answer only $65$--$70\%$ of questions correctly, and its effect is strongly backbone-dependent: Llama-3.3-70B largely ignores the instruction and stays at the baseline (Abs.\ Gap $0.277$ vs.\ $0.279$), whereas Gemini-3.5-Flash-Lite complies and improves across all metrics---but still trails CBUS-SPR on every student-centric metric and CBUS-TS on Pearson correlation. Neither route matches the fidelity of the cognitively bounded simulators. Second, within CBUS-SPR the capacity \emph{bound} is the primary driver of the improvement: \emph{Random dropout} performs on par with CBUS-SPR---better on some metrics and worse on others depending on the backbone---indicating that \emph{how many} propositions are retained matters far more than \emph{which} ones the model selects.

\section{Category-Specific Results}
\label{app:results}

Table~\ref{tab:category_results} breaks down evaluation results per reading comprehension category for two selected LLM backbones, Llama-3.3-70b-Instruct and Gemini-3.5-Flash-Lite. The persona-prompted baseline is markedly less faithful on \emph{Locate} questions than on \emph{Interpret \& Reflect}: baseline JSD is $0.64$--$0.70$ for \emph{Locate} versus $\sim\!0.28$ for \emph{Interpret \& Reflect}, with lower item-difficulty correlations as well, because the superhuman models answer the easier \emph{Locate} items almost uniformly correctly---erasing the natural spread---while making more human-like errors on the harder inferential items. CBUS narrows this gap in both categories but along different axes: the largest student-centric gains come on \emph{Locate} (SPR cuts JSD to $\sim\!0.08$), whereas the strongest item-difficulty correlations are obtained on \emph{Interpret \& Reflect} under TS (Pearson up to $0.556$ and $0.594$).

\begin{table*}[!thbp]
  \caption{Evaluation results broken down by reading comprehension question category, under independent per-student sampling, reported as mean $\pm$ standard deviation over $n=3$ runs. \emph{Interpret \& Reflect} combines the \emph{Interpret \& Connect} and \emph{Reflect \& Evaluate} categories.}
  \label{tab:category_results}
  \centering
  \scriptsize
  \begin{tabular}{lll r@{\,}l r@{\,}l r@{\,}l r@{\,}l r@{\,}l}
    \toprule
                      &                        &                  & \multicolumn{6}{c}{\textbf{Student-Centric}}         & \multicolumn{4}{c}{\textbf{Item-Centric}}                                                                                                                                                                                                                                            \\
    \cmidrule(lr){4-9} \cmidrule(lr){10-13}
    \textbf{Category} & \textbf{Backbone}      & \textbf{Variant} & \multicolumn{2}{c}{\textbf{Abs. Gap} ($\downarrow$)} & \multicolumn{2}{c}{\textbf{JSD} ($\downarrow$)} & \multicolumn{2}{c}{\textbf{ECE} ($\downarrow$)} & \multicolumn{2}{c}{\textbf{Pearson $\rho$} ($\uparrow$)} & \multicolumn{2}{c}{\textbf{Spearman $r_s$} ($\uparrow$)}                                                              \\
    \midrule
    \emph{Locate}
                      & Llama-3.3-70b-Instruct & baseline         & 0.277                                                & \stdev{.001}                                    & 0.640                                           & \stdev{.011}                                             & 0.283                                                    & \stdev{.001} & 0.225 & \stdev{.006} & 0.147 & \stdev{.043} \\
                      &                        & CBUS-TS          & 0.244                                                & \stdev{.001}                                    & 0.470                                           & \stdev{.007}                                             & 0.251                                                    & \stdev{.001} & 0.422 & \stdev{.006} & 0.265 & \stdev{.011} \\
                      &                        & CBUS-SPR         & 0.020                                                & \stdev{.001}                                    & 0.078                                           & \stdev{.002}                                             & 0.066                                                    & \stdev{.002} & 0.306 & \stdev{.007} & 0.293 & \stdev{.002} \\
    \cmidrule(lr){2-13}
                      & Gemini-3.5-Flash-Lite  & baseline         & 0.285                                                & \stdev{.000}                                    & 0.695                                           & \stdev{.001}                                             & 0.290                                                    & \stdev{.000} & 0.252 & \stdev{.001} & 0.168 & \stdev{.017} \\
                      &                        & CBUS-TS          & 0.251                                                & \stdev{.001}                                    & 0.505                                           & \stdev{.009}                                             & 0.256                                                    & \stdev{.001} & 0.410 & \stdev{.003} & 0.260 & \stdev{.005} \\
                      &                        & CBUS-SPR         & 0.022                                                & \stdev{.002}                                    & 0.076                                           & \stdev{.005}                                             & 0.076                                                    & \stdev{.001} & 0.290 & \stdev{.001} & 0.272 & \stdev{.007} \\
    \midrule
    \emph{Interpret \& Reflect}
                      & Llama-3.3-70b-Instruct & baseline         & 0.293                                                & \stdev{.001}                                    & 0.279                                           & \stdev{.003}                                             & 0.302                                                    & \stdev{.001} & 0.279 & \stdev{.008} & 0.311 & \stdev{.050} \\
                      &                        & CBUS-TS          & 0.184                                                & \stdev{.002}                                    & 0.109                                           & \stdev{.003}                                             & 0.188                                                    & \stdev{.002} & 0.556 & \stdev{.009} & 0.484 & \stdev{.048} \\
                      &                        & CBUS-SPR         & 0.167                                                & \stdev{.005}                                    & 0.094                                           & \stdev{.002}                                             & 0.181                                                    & \stdev{.006} & 0.393 & \stdev{.011} & 0.402 & \stdev{.026} \\
    \cmidrule(lr){2-13}
                      & Gemini-3.5-Flash-Lite  & baseline         & 0.290                                                & \stdev{.001}                                    & 0.279                                           & \stdev{.006}                                             & 0.299                                                    & \stdev{.000} & 0.405 & \stdev{.003} & 0.429 & \stdev{.033} \\
                      &                        & CBUS-TS          & 0.132                                                & \stdev{.003}                                    & 0.055                                           & \stdev{.004}                                             & 0.141                                                    & \stdev{.002} & 0.594 & \stdev{.002} & 0.512 & \stdev{.023} \\
                      &                        & CBUS-SPR         & 0.101                                                & \stdev{.001}                                    & 0.049                                           & \stdev{.003}                                             & 0.108                                                    & \stdev{.001} & 0.435 & \stdev{.014} & 0.507 & \stdev{.009} \\
    \bottomrule
  \end{tabular}
\end{table*}

\section{Capacity Sensitivity Analysis}
\label{app:capacity}

\begin{figure}[t]
  \centering
  \includegraphics[width=\linewidth]{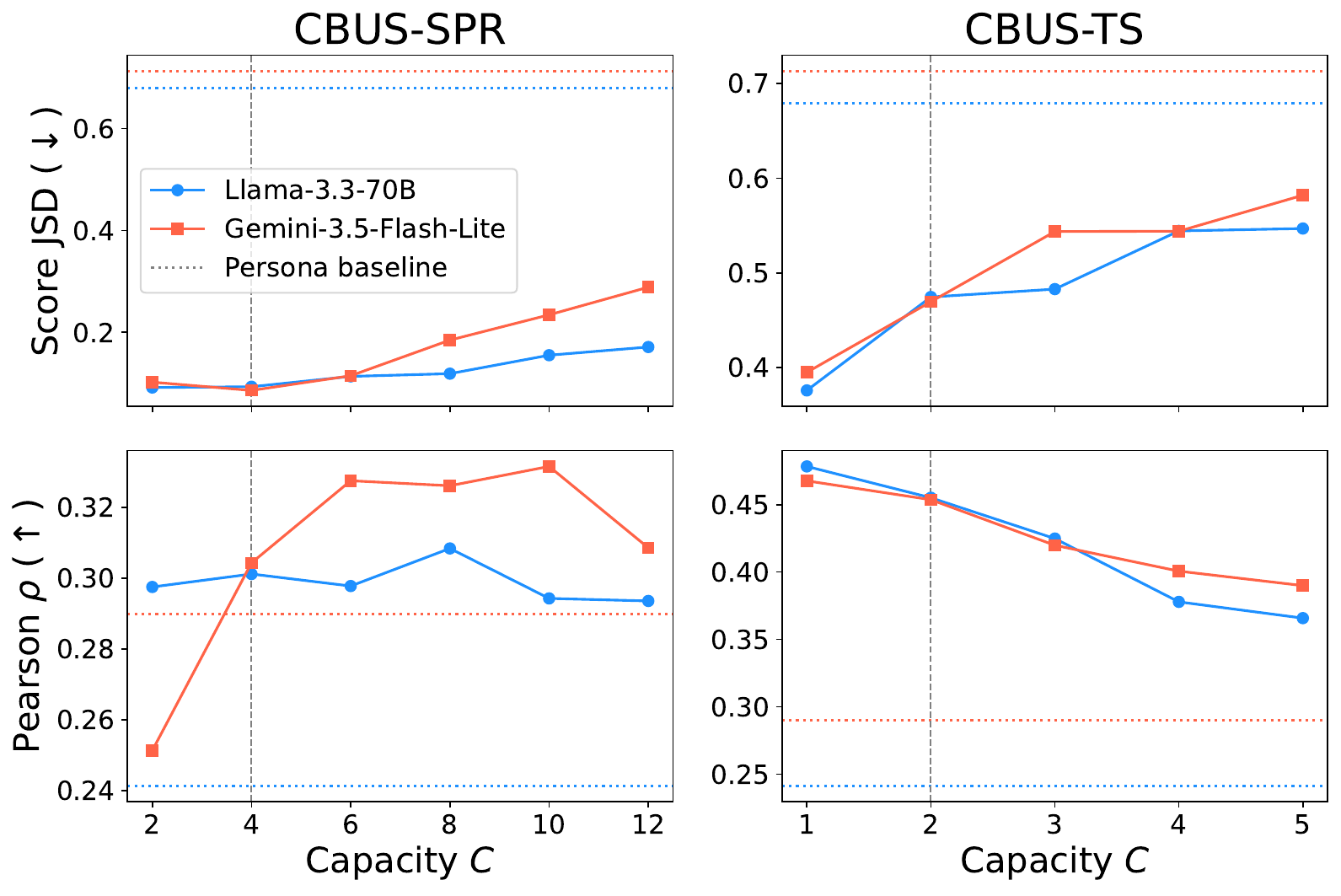}
  \caption{Capacity ($C$) sensitivity for SPR (left) and TS (right) for two selected backbones. Top: student-centric Score JSD ($\downarrow$); bottom: item-centric Pearson $\rho$ ($\uparrow$). Horizontal dotted lines show each backbone's persona-prompted baseline (in the matching color); the vertical dashed line marks the cognitively-motivated operating point (SPR $C=4$, TS $C=2$).}
  \label{fig:capacity_sweep}
\end{figure}

To characterize the effect of the working-memory capacity $C$ and to assess how our cognitively-motivated operating points perform empirically relative to other capacity settings, we perform a full capacity sweep for the two backbones. We vary $C \in \{2,4,6,8,10,12\}$ for SPR and $C \in \{1,2,3,4,5\}$ for TS, and evaluate every setting against the Ground Truth split (Figure~\ref{fig:capacity_sweep}). Given the small run-to-run variance observed in our main experiments (Table~\ref{tab:results}), we report a single run ($n=1$) per capacity setting.

The key pattern is that student-centric and item-centric metrics respond to $C$ in \emph{opposite} directions, so no single capacity is jointly optimal. For SPR, the student-centric error (absolute gap and Score JSD) is lowest at low capacity and grows monotonically as $C$ increases, whereas the item-centric difficulty correlation is flat-to-increasing in $C$: as the reader retains more propositions, its score distribution drifts away from real students even as its item-level discrimination improves. The cognitively-motivated value $C=4$ sits at the student-centric optimum while retaining competitive item-centric alignment. For TS, both families of metrics are best at the smallest capacity ($C=1$) and worsen as $C$ grows; we nonetheless adopt $C=2$, the smallest capacity that still permits integrating information from two points in a text, since $C=1$ cannot, by design, answer questions that require combining multiple facts. Importantly, the approach is robust to the choice of $C$: across the entire sweep, every capacity setting improves over the corresponding baseline on both metrics, the sole exception being CBUS-SPR at $C=2$ for Gemini-3.5-Flash-Lite, whose item-centric correlation dips below its baseline. The qualitative conclusions of Section~\ref{sec:results_and_analysis} are therefore stable across a wide capacity range.

\end{document}